\documentclass{sig-alternate-05-2015}
\usepackage{hyperref}
\usepackage{graphicx}
\usepackage{url}
\usepackage{cite}
\usepackage{amssymb}
\usepackage{amsmath}
\usepackage{algorithm}
\usepackage{algorithmic}
\usepackage{verbatim}
\usepackage{textcomp}
\usepackage{capt-of}
\usepackage{cleveref}
\usepackage{caption}
\usepackage{subcaption}
\usepackage{multirow}
\usepackage{booktabs}
\usepackage{changepage}
\usepackage{afterpage}
\usepackage{xcolor}

\let\OldS\S
\renewcommand{\S}{\OldS{}}

\DeclareMathOperator*{\argmax}{arg\,max}

\begin{document}

\setcopyright{acmcopyright}

\title{Towards Geo-Distributed Machine Learning}

\numberofauthors{5}

\author{
%
%
\alignauthor 
Ignacio Cano\\
       \affaddr{Computer Science \& Engineering}\\
       \affaddr{University of Washington}\\
       \email{icano@cs.washington.edu}
\alignauthor 
Markus Weimer\\
       \affaddr{Cloud and Information Services Lab}\\
       \affaddr{Microsoft}\\
       \email{mweimer@microsoft.com}
\alignauthor 
Dhruv Mahajan\\
       \affaddr{Cloud and Information Services Lab}\\
       \affaddr{Microsoft}\\
       \email{dhrumaha@microsoft.com}
\and  
\alignauthor 
Carlo Curino\\
       \affaddr{Cloud and Information Services Lab}\\
       \affaddr{Microsoft}\\
       \email{ccurino@microsoft.com}
\alignauthor
Giovanni Matteo Fumarola\\
      \affaddr{BigData Team}\\
       \affaddr{Microsoft}\\
       \email{gifuma@microsoft.com}
}

\maketitle

\begin{abstract}

  Latency to end-users and regulatory requirements push large
  companies to build data centers all around the world. The resulting
  data is ``born'' geographically distributed.  On the other hand,
  many machine learning applications require a global view of such
  data in order to achieve the best results.  These types of
  applications form a new class of learning problems, which we call
  Geo-Distributed Machine Learning (GDML). Such applications need to
  cope with: 1) scarce and expensive cross-data center bandwidth, and
  2) growing privacy concerns that are pushing for stricter data
  sovereignty regulations.

  Current solutions to learning from geo-distributed data sources
  revolve around the idea of first centralizing the data in one data
  center, and then training locally. As machine learning algorithms
  are communication-intensive, the cost of centralizing the data is
  thought to be offset by the lower cost of intra-data center
  communication during training.

  In this work, we show that the current centralized practice can be
  far from optimal, and propose a system for doing geo-distributed
  training.  Furthermore, we argue that the geo-distributed approach
  is structurally more amenable to dealing with regulatory
  constraints, as raw data never leaves the source data center.  Our
  empirical evaluation on three real datasets confirms the general
  validity of our approach, and shows that GDML is not only possible
  but also advisable in many scenarios.
  
\end{abstract}

\section{Introduction}



Modern organizations have a planetary footprint. Data is created where
users and systems are located, \emph{all around the globe}.  The
reason for this is two-fold: 1) minimizing latency between serving
infrastructure and end-users, and 2) respecting regulatory
constraints, that might require data about citizens of a nation to
reside within the nation's borders. On the other hand, many machine
learning applications require access to all that data at once to build
accurate models.  For example, fraud prevention systems benefit
tremendously from the global picture in both finance and communication
networks, recommendation systems rely on the maximum breadth of data
to overcome cold start problems, and the predictive maintenance
revolution is only possible because of data from all markets.  These
types of applications that deal with geo-distributed datasets belong
to a new class of learning problems, which we call Geo-Distributed
Machine Learning (GDML).

\begin{figure}
  \centering
  \vspace{2mm}
  \includegraphics[width=1\columnwidth]{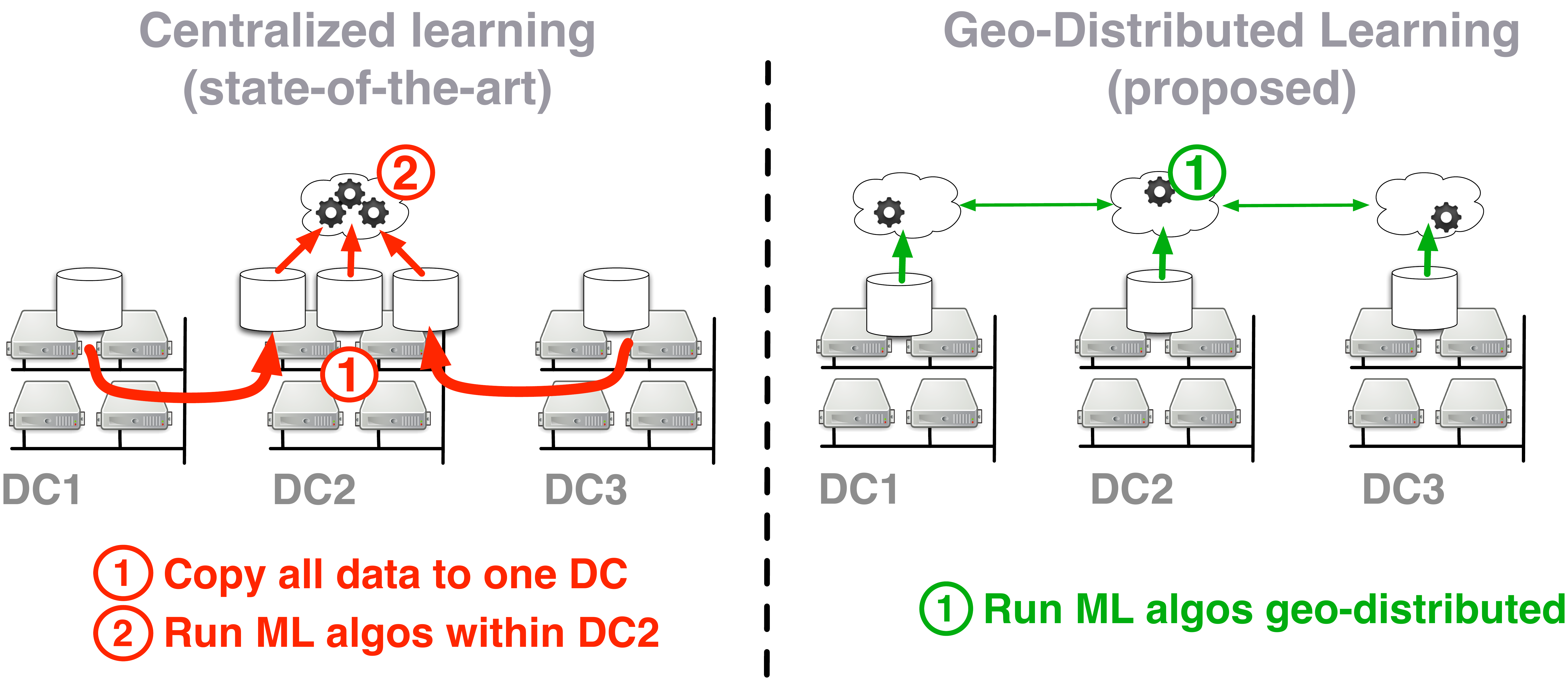}
  \caption{Centralized vs Geo-distributed learning.}
  \label{fig:intro-graphics}
\end{figure}

The state-of-the-art approach to machine learning from decentralized
datasets is to centralize them.  As shown in the left-side of
Figure~\ref{fig:intro-graphics}, this involves a two-step process: 1)
the various partitions of data are copied into a single data center
(DC)---thus recreating the overall dataset in one location, and 2)
learning takes place there, using existing intra-data center
technologies.  Based on conversations with practitioners at Microsoft, 
we gather that this {\em centralized} approach is predominant in most
practical settings. This is consistent with reports on the infrastructures of 
other large organizations, such as Facebook \cite{facebookinfr}, 
Twitter \cite{twitterinfrastructure}, and LinkedIn \cite{auradkar2012data}. 
The reason for its popularity is two-fold, on the one
hand, centralizing the data is the easiest way to reuse existing ML
frameworks~\cite{Zaharia:2012:Spark,Low:2012:GRAPHLAB,Li:2014:PS}),
and on the other hand, machine learning algorithms are notoriously
communication-intensive, and thus assumed to be non-amenable to
cross-data center execution---as we will see, in many practical
settings, we can challenge this assumption.

The centralized approach has two key shortcomings:
%
%

\begin{enumerate}
\item It consumes large amounts of cross-data center (X-DC) bandwidth
  (in order to centralize a copy of the data). X-DC bandwidth has been
  shown to be scarce, expensive, and growing at a slower pace than
  most other intra-data center (in-DC) resources
  \cite{rabkin2014aggregation,Vulimiri:2015:GA,laoutaris2011inter,greenberg2008cost}.
\item It requires raw data to be copied across data centers, thus
  potentially across national borders. While international regulations
  are quickly evolving, the authors of this paper speculate that the growing 
  concerns regarding privacy and data sovereignty \cite{Rost2011,EUPrivacyConcerns}
  might become a key limiting factor to the applicability of centralized learning
  approaches.
\end{enumerate}

We hypothesize that both challenges will persist or grow in the future
\cite{Vulimiri:2015:WAN, SafeHarbor:2015}.

In this paper, we propose the geo-distributed learning approach
(right-side of Figure~\ref{fig:intro-graphics}), where raw data is
kept in place, and learning tasks are executed in a X-DC fashion.  We
show that by leveraging communication-efficient algorithmic solutions
\cite{Dhruv:2013:FADL}, this approach can achieve orders of magnitude
lower X-DC bandwidth consumption in practical settings.  Moreover, as
the distributed-learning approach does not require to copy raw data
outside their native data center (only statistics and estimates are
copied), it is structurally better positioned to deal with evolving
regulatory constraints.  A detailed study of these aspects of the
distributed approach is beyond the scope of this paper.

The solution we propose serves as an initial stand-in for a new
generation of geo-distributed learning systems and algorithms, and
allows us to present the first study on the relative efficiency of
centralized vs. distributed approaches. 
In this paper, we concentrate on two key metrics: X-DC bandwidth
consumption and learning runtime,  and show experimentally that
properly designed centralized solutions can achieve faster learning
times (when the data copy latency is hidden by streaming data as it
becomes available), but that distributed solutions can achieve much
lower X-DC bandwidth utilization, and thus substantially lower cost
for large-scale learning tasks.

Note that while the above two metrics are of great practical relevance,
many other dimensions (e.g. resilience to catastrophic DC failures) 
are worth considering when comparing alternative approaches . 
In \OldS\ref{sec:openproblems}, we briefly list these and other 
open-problems that emerge when considering this new class of learning 
tasks: Geo-Distributed Machine Learning (GDML).

\subsection{Contributions}
Summarizing, our main contributions are:
\begin{itemize}
\item We introduce GDML, an important new class of learning system problems that deals
with geo-distributed datasets, and provide an in-depth study of the relative merits of 
state-of-the-art centralized solutions vs. novel geo-distributed alternatives. 

\item We propose a system that builds upon Apache Hadoop YARN~\cite{Vavilapalli:2013:AHY} 
and Apache REEF \cite{Weimer:2015:REEF}, and extends their functionality to support 
	multi-data center ML applications. We adopt a communication-sparse learning algorithm 
	\cite{Dhruv:2013:FADL}, originally designed to accelerate learning, and leverage it to
	lower the bandwidth consumption (i.e. cost) for geo-distributed learning.
 
\item We present empirical results from both simulations and
a real deployment across continents, which show that under common conditions 
distributed approaches can trade manageable penalties in training latency 
(less than $5\times$) for massive bandwidth reductions (multiple orders of magnitude). 

\item Finally, we highlight that GDML is a tough new challenge, and that many problems, such 
as WAN fault-tolerance, regulatory compliance, privacy preservation, X-DC scheduling coordination, 
latency minimization, and support for broader learning tasks remain open. 
\end{itemize}

The remainder of this paper presents these findings as follows:
\OldS\ref{sec:problem} formalizes the problem setting,
\OldS\ref{sec:approach} introduces our approach, \OldS\ref{sec:algo}
introduces the algorithmic solution, and \OldS\ref{sec:system}
describes our system implementation. We explain the evaluation setup
in \OldS\ref{sec:evaluation} and discuss the experimental results in
\OldS\ref{sec:discussion}.  Finally, we discuss related work in
\OldS\ref{sec:related_work}, open problems in \OldS\ref{sec:openproblems} 
and conclusions in \OldS\ref{sec:conclusions}.

\section{Problem Formulation}\label{sec:problem}

In order to facilitate a study of the state-of-the-art centralized
approach in contrast to potential geo-distributed alternatives, we
formalize the problem below in two dimensions: 1) we specify
assumptions about the data, its size and partitioning, and 2) we
restrict the set of learning problems to the well known Statistical
Query Model (SQM~\cite{Kearns:1998}) class.

\subsection{Data distribution}
We assume the dataset $D$ of $N$ examples $(x_i,y_i)$ to exist in $p
\in \{1,\dots,P\}$ partitions $D_p$, each of which is generated in one
of $P$ data centers. Those $P$ partitions consist of $n_p$ examples
each, with $N=\sum_{p}n_p$. Further, let $d$ be the dimensionality of
the feature vectors and $\bar{d}$ the average sparsity (number of
non-zeros) per example. The total size of each partition can be
estimated as $s_p=n_p \cdot \bar{d}$.

Let $p^*$ be the index of the largest partition, i.e. $p^* =
\argmax_{p} n_p$. Then, the total X-DC transfer needed to centralize
the dataset is:
\begin{equation}
 T_C = (N-n_p)\bar{d}
\label{eq:dtcost}
\end{equation}

Data compression is commonly applied to reduce this size, but only by
a constant factor.

\subsection{Learning Task}

For any meaningful discussion of the relative merits of the
centralized approach when compared to alternatives, we need to
restrict the set of learning problems we consider. Here, we choose
those that fit the Statistical Query Model (SQM)~\cite{Kearns:1998}.
This model covers a wide variety of important practical machine
learning models, including K-Means clustering, linear and logistic
regression, and support vector machines.  In the Statistical Query
Model, the algorithm can be phrased purely in terms of statistical
queries over the data. Crucially, it never requires access to
individual examples, and the statistical queries decompose into the
sum of a function applied to each data point~\cite{NIPS2006_3150}. Let
that query function be denoted by $f_q \in \{f_1, f_2, \dots f_Q\}$. A
query result $F_q$ is then computed as $F_q = \sum_{i=1}^N
f_q(x_i,y_i)$.

With the data partitioning, this can be rephrased as

\begin{equation}
F_q = \sum_{p=1}^{P}\sum_{i=1}^{n_p} f_q(x_i,y_i)
\end{equation}

In other words, the X-DC transfer per statistical query is the size of
the output of its query function $f_q$, which we denote as $s_q$.  The
total X-DC transfer then depends on the queries and the number of such
queries issued, $n_q$, as part of the learning task, both of which
depend on the algorithm executed and the dataset.  Let us assume we
know these for a given algorithm and dataset combination.  Then we
can estimate the total X-DC transfer cost of a fully distributed
execution as:

\begin{equation}
  T_D = P \sum_{q=1}^Q n_q \cdot s_q
	\label{eq:xdcqcost}
\end{equation}

Note that this relies on the cumulative and associative properties of
the query aggregation: we only need to communicate one result of size
$s_q$ per data center.

With this formalization, the current state-of-the-art approach of
centralizing the data relies on the assumption that $T_C <<
T_D$. However, it isn't obvious why this should always be the case:
the X-DC transfer of the centralized approach $T_C$ grows linearly with   
the dataset size, whereas the X-DC transfer of a distributed approach
$T_D$ grows linearly with the size and number of the queries. Hence, it
is apparent that the assumption holds for some, but not all regimes.
All things being equal, larger datasets favor the distributed
approach. Similarly, all things being equal, larger query results and
algorithms issuing more queries favor the centralized approach.

In order to study this more precisely, we need to restrict the
discussion to a concrete learning problem for which the queries $q$,
their functions $f_q$ and their output sizes $s_q$ are known. Further,
the number of such queries can be bounded by invoking the convergence
theorems for the chosen learning algorithm. Here, we choose linear
modeling to be the learning problem for its simplicity and rich
theory.  In particular, we consider the $l_2$ regularized linear
learning problem. 

Let $l(w \cdot x_i, y_i)$ be a continuously differentiable loss
function with Lipschitz continuous gradient, where $w \in
\mathbb{R}^d$ is the weight vector. Let $L_p(w) = \sum_{i \in D_p} l(w
\cdot x_i, y_i)$ be the loss associated with data center $p$, and
$L(w) = \sum_{p} L_p(w)$ be the total loss over all data centers.  Our
goal is to find $w$ that minimizes the following objective function,
which decomposes per data center:
\begin{eqnarray}
  f(w) = \frac{\lambda}{2} ||w||^2 + L(w) = \frac{\lambda}{2} ||w||^2 + \sum_{p} L_p(w) \label{eq:l2}
\end{eqnarray}
where $\lambda > 0$ is the regularization constant. Depending on the
loss chosen, this objective function covers important cases such as
linear support vector machines and logistic regression. Learning such
model amounts to optimizing~\eqref{eq:l2}.  Many optimization
algorithms are available for the task, and in \OldS\ref{sec:algo} we
describe the one we choose.

It is important to note that one common statistical query of all those
algorithms is the computation of the gradient of the model
in~\eqref{eq:l2} with respect to $w$. The size of that gradient (per
partition and globally) is $d$. Hence, $s_q$ for this class of models
can be approximated by $d$. This allows us to rephrase the trade-off
mentioned above. All things being equal, datasets with more examples
$(x_i,y_i)$ favor the distributed approach. Similarly, all things
being equal, datasets with wider dimensionality $d$ favor the
centralized approach.

\section{Approach}\label{sec:approach}

In order to study the questions raised above, we need two major
artifacts: 1) a communication-efficient algorithm to
optimize~\eqref{eq:l2}, and 2) an actual geo-distributed
implementation of that algorithm.  In this section, we describe both
of these items in detail.

\subsection{Algorithm}\label{sec:algo}

As mentioned above, we focus on the X-DC communication costs
here. Hence, we need a communication-efficient algorithm capable of
minimizing the communication between the data centers. It is clear
from~\eqref{eq:xdcqcost} that such an algorithm should try to minimize
the number of queries whose output size is very large. In the case
of~\eqref{eq:l2}, this means that the number of X-DC gradient computations
should be reduced.

Recently, many communication-efficient algorithms have been proposed
that trade-off local computation with communication
\cite{JMLR:v15:agarwal14a,Kearns:1998,Boyd:2011:DOS:2185815.2185816,
  DBLP:journals/jmlr/ZhangLS12,DBLP:conf/nips/JaggiSTTKHJ14}. In this
work, we use the algorithm proposed by Mahajan et
al.~\cite{Dhruv:2015:FADL} to optimize~\eqref{eq:l2}, shown in
Algorithm \ref{fadl-algorithm}. We choose this algorithm because
experiments show that this method performs better than the
aforementioned cited ones, both in terms of communication passes and
running time \cite{Dhruv:2015:FADL}. The algorithm was initially
thought for running in a traditional distributed machine learning
setting, i.e. single data center\footnote{We confirmed this with the
  first author as per 2/2016.}. In this work, we adapt it for X-DC
training, a novel application that wasn't originally intended for.

The main idea of the algorithm is to trade-off in-DC computation and
communication with X-DC communication.  The minimization of the
objective function $f(w)$ is performed using an iterative descent
method in which the $r$-th iteration starts from a point $w^r$,
computes a direction $d^r$, and then performs a line search along that
direction to find the next point $w^{r+1} = w^r + t\,d^r$.

We adapt the algorithm to support GDML in the following manner. Each node in
the algorithm now becomes a data center. All the local computations
like gradients and loss function on local data now involves both
computation as well as in-DC communication among the nodes in the same
data center.  On the other hand, all global computations like gradient
aggregation involves X-DC communication. This introduces the need for
two levels of communication and control, described in more detail in 
\OldS~\ref{sec:system}.

In a departure from communications-heavy methods, this algorithm uses
distributed computation for generating a good search direction $d^r$
in addition to the gradient $g^r$. At iteration $r$, each data center
has the current global weight vector $w^r$ and the gradient
$g^r$. Using its local data $D_p$, each data center can form an
approximation $\hat{f}_p$ of $f$.  To ensure convergence, $\hat{f}_p$
should satisfy a gradient consistency condition, $\nabla
\hat{f}_p(w^r) = g^r$.  The function $\hat{f}_p$ is
approximately\footnote{Mahajan et. al~\cite{Dhruv:2015:FADL} proved
  linear convergence of the algorithm even when the local problems are
  optimized approximately.}  optimized using a method $M$ to get the
local weight vector $w_p$, which enables the computation of the local
direction $d_p = w_p - w^r$. The global update direction is chosen to
be $d^r = \frac{1}{P} \sum_{p} d_p$, followed by a line search to find
$w^{r+1}$.

In each iteration, the computation of the gradient $g^r$ and the
direction $d^r$ requires communication across data centers.  Since
each data center has the global approximate view of the full objective
function, the number of iterations required are significantly less
than traditional methods, resulting in orders of magnitude
improvements in terms of X-DC communication.

The algorithm offers great flexibility in choosing $\hat{f}_p$ and the
method $M$ used to optimize it. A general form of $\hat{f}_p$
for~\eqref{eq:l2} is given by:

\begin{equation}
  \hat{f}_p(w) = \frac{\lambda}{2} ||w||^2 + \tilde{L}_p(w) + \hat{L}_p(w)
  \label{eq:fphat}
\end{equation}

where $\tilde{L}_p$ is an approximation of the total loss $L_p$
associated with data center $p$, and $\hat{L}_p(w)$ is an
approximation of the loss across all data centers except $p$,
i.e. $L(w) - L_p(w) = \sum_{q \neq p} L_q(w)$.  Among the possible
choices suggested in \cite{Dhruv:2015:FADL}, we consider the following
quadratic approximations\footnote{One can simply use $\tilde{L}_p =
  L_p$, i.e. the exact loss function for data center p. However,
  Mahajan et. al~\cite{Dhruv:2015:FADL} showed better results if the
  local loss function is also approximated.} in this work:

\begin{eqnarray}
  \tilde{L}_p(w) &=& \nabla L_p(w^r)(w - w^r)\\
  & & + \frac{1}{2} (w-w^r)^T H_p^r (w-w^r) \label{eq:lp} \nonumber \\
  \hat{L}_p(w) &=& (\nabla L(w^r) - \nabla L_p(w^r)) (w - w^r) \\ 
  & & + \frac{P-1}{2} (w-w^r)^T H_p^r (w - w^r) \label{eq:lp_hat}\nonumber 
\end{eqnarray}
where $H_p^r$ is the Hessian of $L_p$ at $w^r$.\\
Replacing (6)~$+$~(7) in \eqref{eq:fphat} we
have the following objective function:
\begin{eqnarray}
  \hat{f}_p(w) &=& \frac{\lambda}{2} ||w||^2 + g^r \cdot (w - w^r)  \\
  && + \frac{P}{2} (w-w^r)^T H_p^r (w - w^r) \nonumber \label{eq:new_fp_hat}
\label{eq:quadapprox}
\end{eqnarray}

We use the conjugate gradient (CG) algorithm~\cite{CG} to
optimize~(8).  Note that each iteration of CG
involves a statistical query with output size $d$ to do a
hessian-vector computation.  However, this query involves only in-DC
communication and computation, whereas for traditional second order
methods like TRON~\cite{Lin:2008:TRN}, it will involve X-DC
communication.

{\bf{Discussion: }} Let $T_{outer}$ be the number of iterations
required by the algorithm to converge.  Each iteration requires two
queries with output size $s_q=d$ for the gradient and direction
computation, and few queries of output size $s_q=1$ for the objective
function computation in the line search.  Since $d>>1$, we can ignore the
X-DC communication cost for the objective function computation. Hence, we
can rewrite~\eqref{eq:xdcqcost} as $T_D = 2PdT_{outer}$. 
Hence, for $T_D$ to be less than $T_C$ the following
must hold:

\begin{eqnarray}
  2PdT_{outer} &< (N-n_p)\bar{d}
\end{eqnarray}

In practice, the typical value of $P$ (data centers) is relatively
small (in the 10s). Since there are few data centers (i.e. few partitions of the data), 
the above algorithm will take only few (5-20) outer
iterations to converge. In fact, in all our experiments in
\OldS\ref{sec:data}, the algorithm converges in less than $7$ iterations. This
means that as long as the total size of the data is roughly more than $2-3$
orders of magnitude greater than the dimensionality $d$, it is always
better to do geo-distributed learning. Note that for large datasets
this is typically the case and hence the proposed approach should work
better.

\begin{algorithm}[tb]
\begin{algorithmic}
  \STATE{Choose $w^0$}
  \FOR{$r = 0,1 ... $}
     \STATE{Compute $g^r$ (X-DC communication)}
     \STATE{Exit if $||g^r|| \leq \epsilon_g ||g^0||$}
     \FOR{$p=1,...,P$ (in parallel)}
      \STATE Construct $\hat{f}_p(w)$ ((8))
			\STATE $w_p \gets$ Optimize $\hat{f}_p(w)$ (in-DC communication)
     \ENDFOR
     \STATE{$d^r \gets \frac{1}{P} \sum_{p} w_p - w^r$} (X-DC communication)
     \STATE{Line Search to find $t$ (negligible X-DC communication)\\
     $w^{r+1} \gets w^r + t\,d^r$}
   \ENDFOR
\end{algorithmic}
\caption{Functional Approximation based Distributed Learning Algorithm (FADL)}
\label{fadl-algorithm}
\end{algorithm}

\subsection{Distributed Implementation}\label{sec:system}

The algorithm above requires the flexibility to be executed in an intra-data center 
environment as well as in a X-DC one.  We need a system that
can run in these two regimes without requiring two separate
implementations.  Here, we describe such flexible system.

The algorithm introduced in \OldS\ref{sec:algo} relies on Broadcast and Reduce operators.  
As part of this work, we add X-DC versions of those to Apache REEF (\OldS\ref{sec:reef}), 
which provides the basic control flow for our implementation. 
Moreover, our system needs to
obtain resources (CPU cores, RAM) across different data centers in a uniformly basis.
Such resources are managed by a resource manager in our current
architecture. Further, Apache Hadoop YARN's 
new federation feature (\OldS\ref{sec:rm}) allows us to \emph{view} 
multiple data centers as a single one. 
We extend Apache REEF with support for that. 
Below, we provide more details on our three-layer architecture, from bottom to top\footnote{All changes to REEF and YARN have been contributed back to the projects where appropriate.}.

\subsubsection{Resource Manager: Apache Hadoop YARN}
\label{sec:rm}
A resource manager (RM) is a platform that dynamically leases
resources, known as containers, to various competing applications in a
cluster.  It acts as a central authority and negotiates with
potentially many Application Masters (AM) the access to those
containers \cite{Weimer:2015:REEF}.  The most well known
implementations are Apache Hadoop YARN \cite{Vavilapalli:2013:AHY},
Apache Mesos \cite{Hindman:2011:MPF} and Google Omega
\cite{Schwarzkopf:2103:SIGOPS}. All of these systems are designed to
operate {\em within} one DC and multiplex applications on a collection
of shared machines.

In our setting, we need a similar abstraction, but it must \emph{span}
multiple DCs.  We build our solution on top of Apache Hadoop YARN. As
part of our effort to scale-out YARN to Microsoft-scale clusters (tens
of thousands of nodes), we have been contributing to Apache a new
architecture that allows to \emph{federate} multiple clusters \cite{yarnfederationjira}.  
This effort was not originally intended to operate in a X-DC setting, and as such, 
was focused on \emph{hiding} from the application layer the different
sub-clusters.  As part of this work, we have been experimenting and
extending this system, leveraging its transparency yet providing
enough visibility of the network topology
to our application layer (REEF).  
As a result, we can run a \emph{single} application that
spans different data centers in an efficient manner.

\subsubsection{Control Flow: Apache REEF}
\label{sec:reef}
Apache REEF~\cite{Weimer:2015:REEF} provides a generalized control
plane to ease the development of applications on resource managers.
REEF provides a control flow master called Driver to applications, and
an execution environment for tasks on containers called Evaluator.
Applications are expressed as event handlers for the Driver to perform
task scheduling (including fault handling) and the task code to be
executed in the Evaluators. As part of this work, we extend REEF to
support geo-federated YARN, including scheduling of resources to
particular data centers.

REEF provides a group communications library that exposes Broadcast
and Reduce operators similar to Message Passing Interface (MPI)
\cite{MPI:98}.  REEF's group communications library is designed for
the single data center case. We expand it to cover the X-DC case we
study here.

\begin{figure}
  \centering
  \includegraphics[width=1\columnwidth]{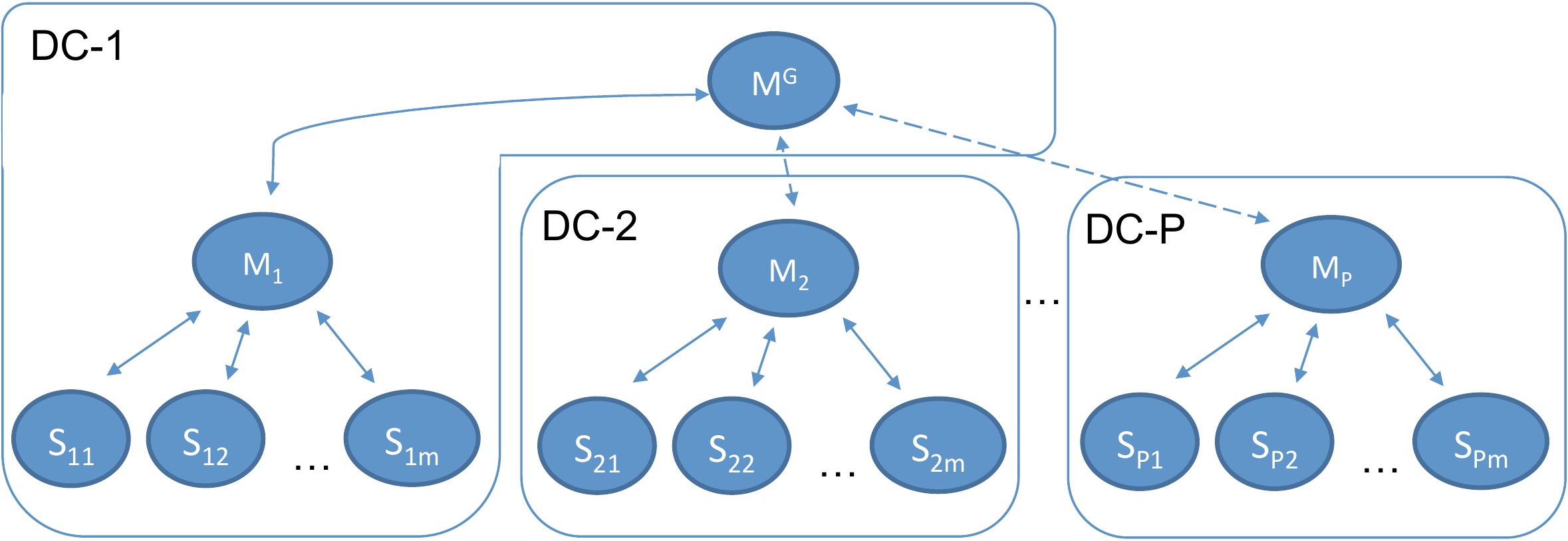}
  \caption{Multi-Level Master/Slave Communication Tree with $P$ data
    centers, each with its own data center master ($M_i$) and slaves
    ($S_{ij}$). The global master $M^G$ is physically located in
    DC-1. The solid and dashed lines refer to in-DC and X-DC links
    respectively. Our current implementation supports recovery from
    slave failures. Fault-tolerance at the master levels is left for
    future work.}
  \label{tree}
\end{figure}

\subsubsection{GDML Framework}\label{sec:trees}

Statistical Query Model (SQM~\cite{Kearns:1998}) algorithms, such as
the one introduced in \OldS\ref{sec:algo}, can be implemented using
nothing more than Broadcast and Reduce operators \cite{NIPS2006_3150},
where data partitions reside in each machine, and the statistical
query is Broadcast to those, while its result is computed on each
machine and then aggregated via Reduce.

Both Broadcast and Reduce operations are usually implemented via
communication trees in order to maximize the overall throughput of the
operation. Traditional systems such as MPI~\cite{MPI:98}
implementations derive much of their efficiency from intelligent (and
fast) ways to establish these trees.  Different from the in-DC
environment where those are typically used, our system needs to work
with network links of vastly different characteristics. X-DC links
have higher latency than in-DC ones, whereas the latter have usually
higher bandwidths \cite{Ballani:2011:TPD:2018436.2018465}.  Further,
X-DC links are much more expensive than in-DC links, as they are
frequently rented or charged-for separately, in the case of the public
cloud.

In our system, we address these challenges with a heterogeneous,
multi-level communication tree.

Figure~\ref{tree} shows an example of the multi-level communication
tree we use.  A global Broadcast originates from $M^G$ to the data
center masters $M_i$, which in turn do a local Broadcast to the slave
nodes $S_{ij}$ in their own data centers. Conversely, local Reduce
operations originate on those slave nodes, while the data center
masters $M_i$ aggregate the data prior to sending it to $M^G$ for
global aggregation.

\begin{figure}
  \centering
  \includegraphics[width=1\columnwidth]{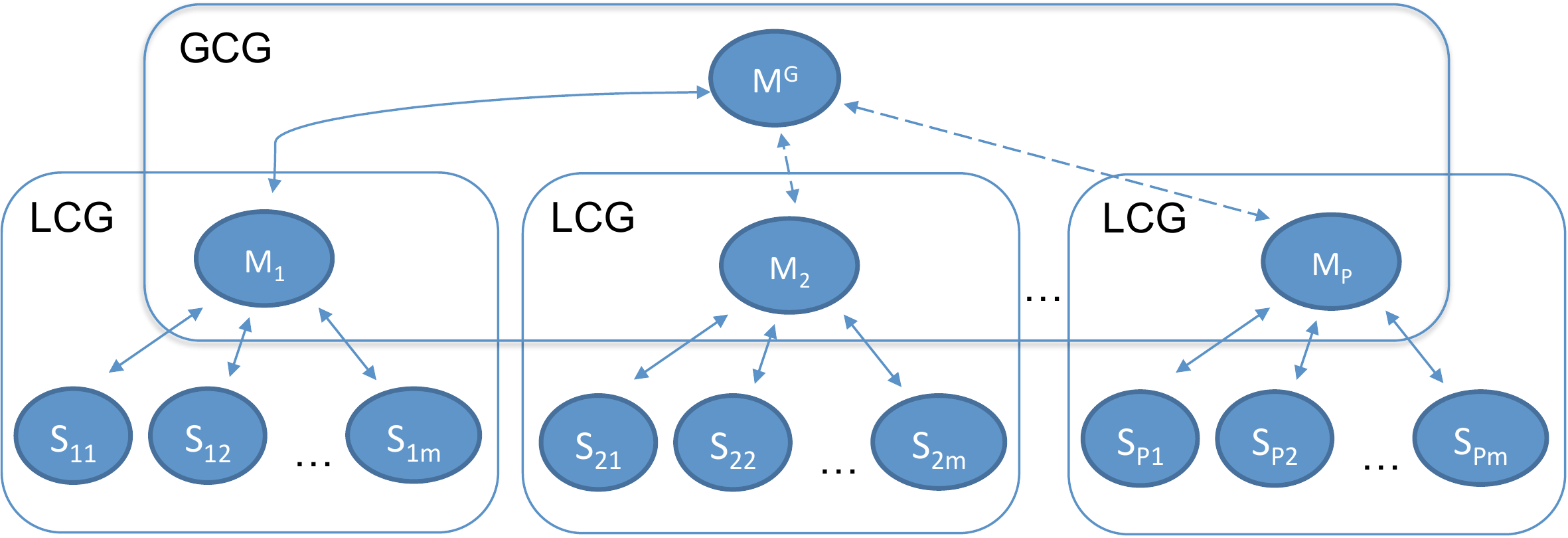}
  \caption{Communication Groups view of the Multi-Level Master/Slave
    Communication Tree depicted in Figure \ref{tree}. There are $P$
    local communication groups (LCG) that connect the data center
    masters ($M_i$) with their respective slaves ($S_{ij}$) to enable
    local Broadcast / Reduce operations. Further, a
    single global communication group (GCG) enables the interaction
    between the global master $M^G$ and the data center masters
    $M_i$ for global aggregation.}
  \label{comm-groups}
\end{figure}

To make this happen, the underlying implementation creates multiple
communication groups, as shown in Figure~\ref{comm-groups}.  The
global master $M^G$ together with the data centers masters $M_i$, form
the global communication group (GCG), where the global Broadcast /
Reduce operations are performed, and used in the outer loop of
Algorithm \ref{fadl-algorithm}.  Likewise, the slave nodes within each
data center and their masters $M_i$ form the local communication
groups (LCG), where the local Broadcast / Reduce operations execute,
and are used to optimize the local approximations $\hat{f_p}$ of 
(8).

\section{Evaluation}\label{sec:evaluation}

The algorithm and system presented above allow us to evaluate the
state-of-the-art of centralizing the data before learning in
comparison with truly distributed approaches. In this section, we
describe our findings, starting with the setup and definition of the
different approaches used, followed by experimental results from both
simulated and real deployments.

\subsection{Experimental Setup}

We report experiments on two deployments: a distributed deployment on
Microsoft~Azure across two data centers, and a large centralized
cluster on which we simulate a multi-data center setup (2, 4, and 8
data centers). This simulated environment is our main test bench, and
we mainly use it for multi-terabyte scale experiments, which aren't
cost-effective on public clouds. We use 256 slave nodes divided
into 2-8 simulated datacenters in all our simulations.
Further, all the experiments are done with the logistic loss function.

We ground and validate the findings from the simulations on a real
cross-continental deployment on Microsoft Azure.  We establish two
clusters, one in a data center in Europe and the other on the
U.S. west coast.  We deploy two DS12 VMs into each of these
clusters.  Each of those VMs has 4 CPU cores and 28GB of RAM.  We
establish the site-to-site connectivity through a VPN tunnel using a
High Performance VPN
Gateway\footnote{\url{https://azure.microsoft.com/en-us/documentation/articles/vpn-gateway-about-vpngateways/}}.

\subsection{Data}\label{sec:data}

User behavior data of web sites is one of the most prominent cases
where data is born distributed: the same website or collection of
websites is deployed into many data centers around the world to
satisfy the latency demands of its users.  Hence, we choose three datasets 
from this domain for our evaluation, two of which are publicly
available.  All of them are derived from click logs.
Table~\ref{datasettable} summarizes their statistics.  CRITEO and
KAGGLE are publicly available~\cite{Criteo,Kaggle}.  The latter is a
small subset of the former, and we use it for the smaller scale
experiments in Azure.  WBCTR is an internal Microsoft dataset.  We
vary the number of features in our experiments using hashing kernels
as suggested in~\cite{Weinberger2009}.

The dataset sizes reported in Table~\ref{datasettable} refer to
compressed data.  The compression/decompression is done using
Snappy\footnote{\url{https://github.com/xerial/snappy-java}}, which
enables high-speed compression and decompression with reasonable
compression size.  In particular, we achieve compression ratios of
around 62-65\% for the CRITEO and WBCTR datasets, and 50\% for KAGGLE.
Following current practice in large scale machine learning, our system
performs all computations using double precision arithmetic, but
communicates single precision floats. Hence, model sizes in
Table~\ref{datasettable} are reported based on single-precision
floating point numbers.

\begin{table}[t!]
\begin{center}
\begin{tabular}{lccccc}
\toprule

\multirow{2}{*}{\textbf{Dataset}} & \textbf{Examples} 
& \textbf{Features} & \multicolumn{2}{c}{\textbf{Size}}
\\
& \textbf{(N)} & \textbf{(d)} & \textbf{Model} & \textbf{Dataset} \\
\midrule
\multirow{4}{*}{CRITEO} & \multirow{4}{*}{4B} & 5M & 20MB & 1.5TB  \\
  & & 10M & 40MB & 1.5TB  \\
  & & 50M & 200MB & 1.6TB  \\
  & & 100M & 400MB & 1.7TB  \\
\midrule
\multirow{4}{*}{WBCTR} & \multirow{4}{*}{730M} & 8M & 32MB & 347GB  \\
  & & 16M & 64MB & 362GB  \\
  & & 80M & 320MB & 364GB  \\
  & & 160M & 640MB & 472GB  \\
\midrule
\multirow{4}{*}{KAGGLE} & \multirow{4}{*}{46M} & 0.5M & 2MB & 8.5GB \\
  & & 1M & 4MB & 8.5GB  \\
  & & 5M & 20MB & 9GB  \\
\bottomrule
\end{tabular}
\end{center}
\caption{Datasets statistics. Dataset sizes reported are \emph{after}
  compression. Weights in the models are represented in
  single-precision floating-point format (32 bits) with no further compression.}
\label{datasettable}
\end{table}



\subsection{Methods}\label{sec:methods}

We contrast the state-of-the-art approach of centralizing the data
prior to learning with several alternatives, both within the regime
requiring data copies and truly distributed:
\begin{description}
\item[\emph{centralized}] denotes the current state-of-the art, where we copy
  the data to one data center prior to training.
  Based on the data shipping model used, two variants of this
  approach arise:  
  \begin{description}
  \item[\emph{centralized-stream}] refers to a streaming copy model where the
    data is replicated as it arrives. When the learning job is
    triggered in a particular data center, the data has already been
    transferred there, therefore, no copy time is included in the job
    running time, and
  \item [\emph{centralized-bulk}] refers to a batch replication scheme
    where the data still needs to be copied by the time the learning
    process starts, therefore, the copy time has an impact on the job
    running time, i.e. the job needs to wait until the transfer is
    made to begin the optimization.
  \end{description}
  We observe both flavors occur in practice.  We simply refer to
  \emph{centralized} when no distinction between its variants is
  required.  This approach (and its variants) only performs
  \emph{compressed} data transfers, and uses the algorithm described
  in \OldS\ref{sec:algo} for solving the $l_2$ regularized linear
  classification problem mentioned in \OldS\ref{sec:problem}.
\item[\emph{distributed-fadl}] also uses the algorithm introduced in
  \OldS\ref{sec:algo} to optimize~\eqref{eq:l2}, but performs
  the optimization in a geo-distributed fashion, i.e.  it leaves the
  data in place and runs a single job that spans training across data
  centers.
\item[\emph{distributed}] builds the multi-level master/slave tree for
  X-DC learning, but does not use the communication-efficient
  algorithm in \OldS\ref{sec:algo} to optimize~\eqref{eq:l2},
  instead, it optimizes using TRON \cite{Lin:2008:TRN}.
\end{description}

Both \emph{distributed} and \emph{distributed-fadl} methods represent
the furthest departure from the current state-of-the-art as their
execution is truly geo-distributed.  Studying results from both allows
us to draw conclusions about the relative merits of a
communications-sparse algorithm and a system enabling truly
geo-distributed training.

\begin{figure*}[tbp!]
    \centering
    \begin{subfigure}{0.24\textwidth}
        \includegraphics[width=\textwidth]{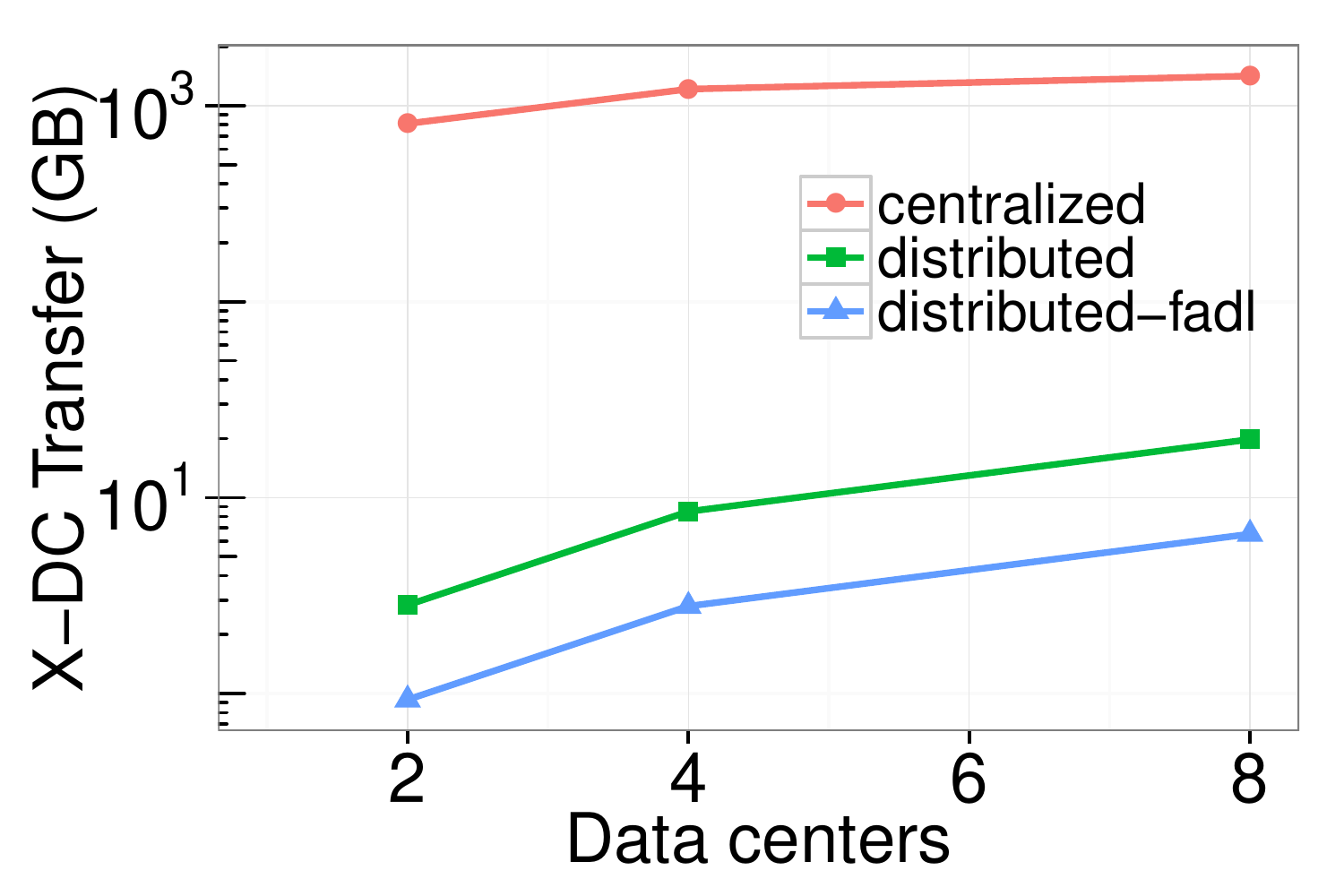}
        \caption{CRITEO 10M}
        \label{fig:criteo-transfer-datacenter-10M}
    \end{subfigure}
    \begin{subfigure}{0.24\textwidth}
        \includegraphics[width=\textwidth]{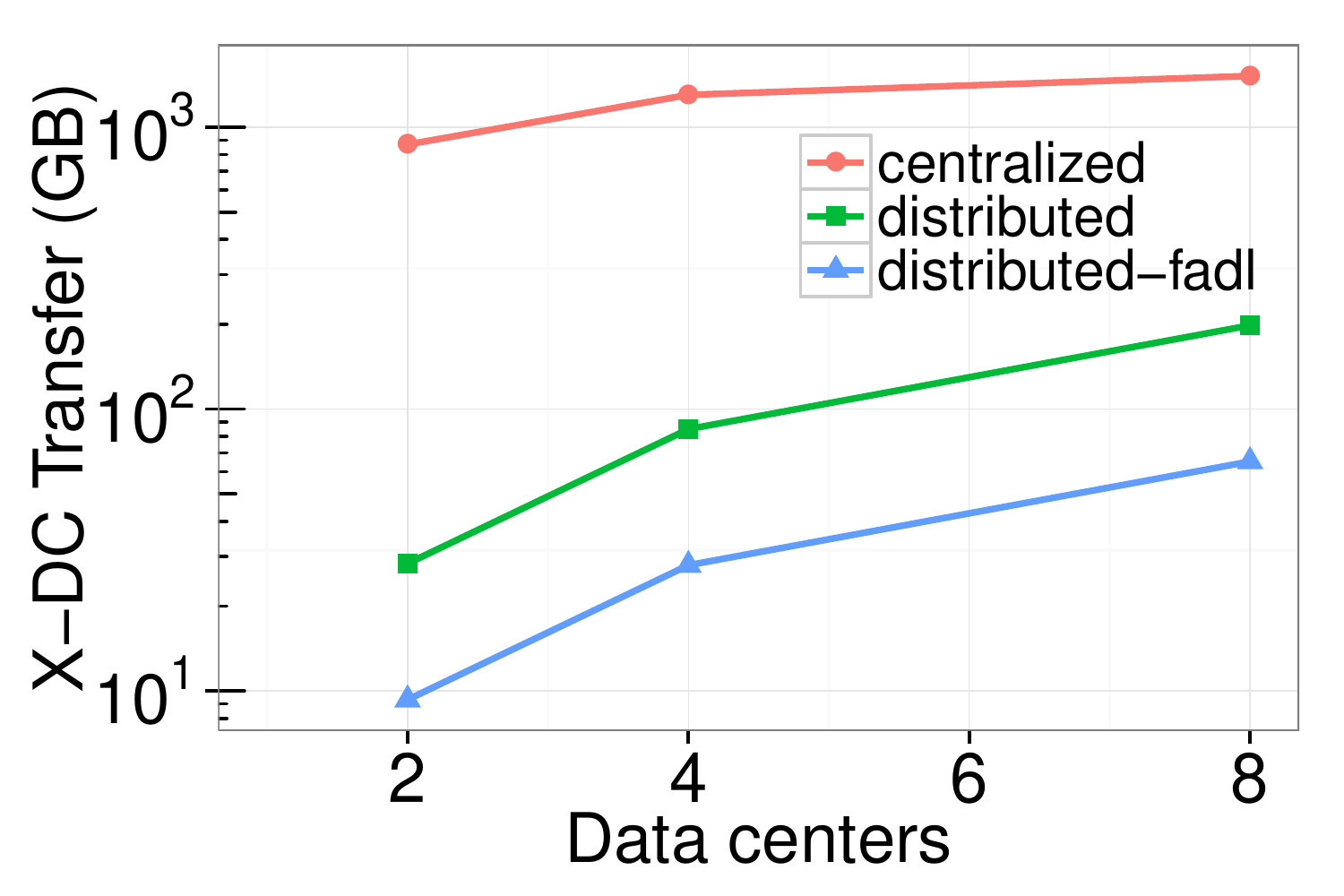}
        \caption{CRITEO 100M}
        \label{fig:criteo-transfer-datacenter-100M}
    \end{subfigure}
    \begin{subfigure}{0.24\textwidth}
        \includegraphics[width=\textwidth]{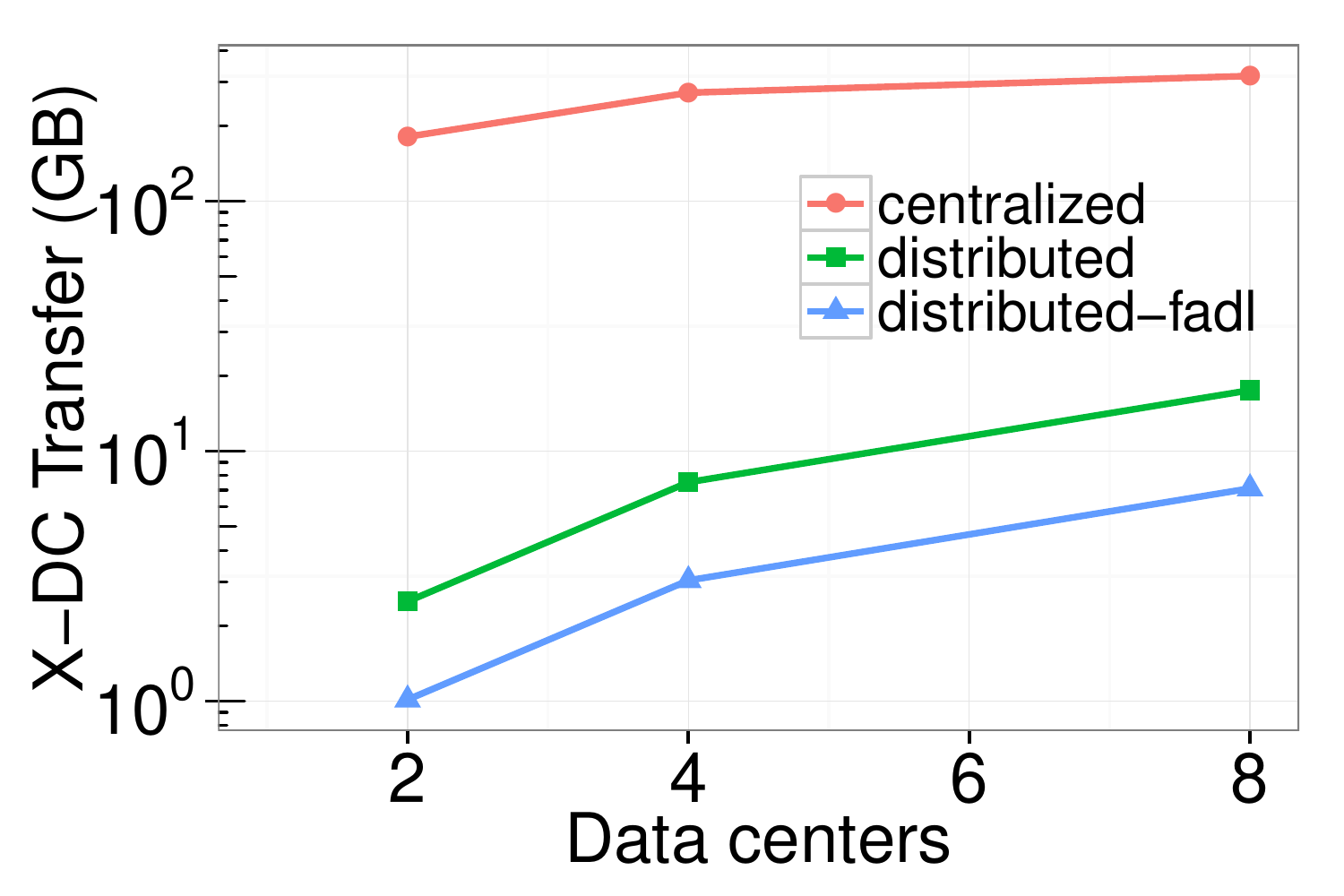}
        \caption{WBCTR 16M}
        \label{fig:wood-transfer-datacenter-16M}
    \end{subfigure}
    \begin{subfigure}{0.24\textwidth}
        \includegraphics[width=\textwidth]{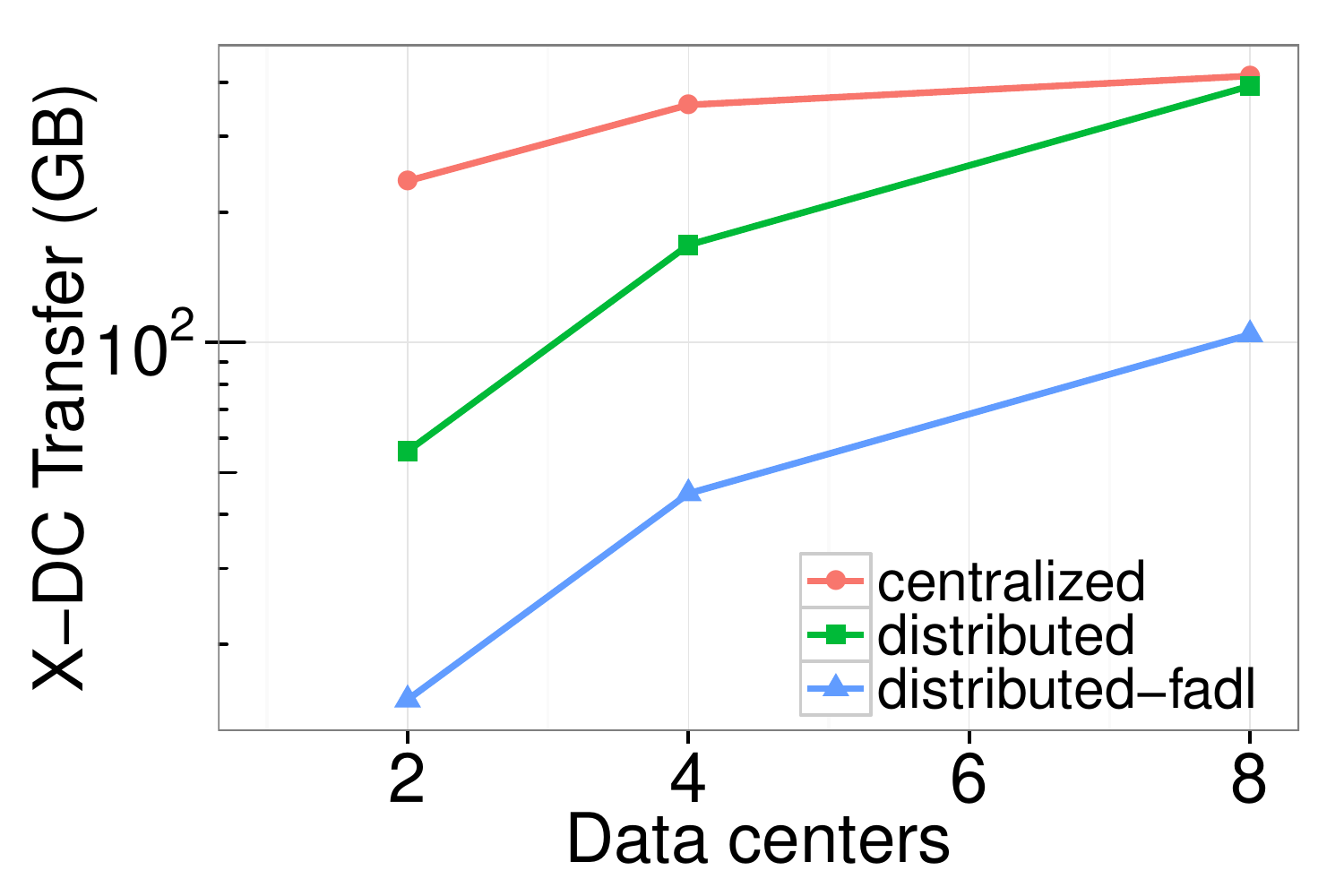}
        \caption{WBCTR 160M} 
        \label{fig:wood-transfer-datacenter-160M}
    \end{subfigure}
    \caption{X-DC transfer (in GB) versus number of data centers for
      two versions of CRITEO and WBCTR datasets. The method
      \emph{distributed-fadl} consumes orders of magnitude less
      X-DC bandwidth than any variant (\emph{stream} or \emph{bulk}) 
      of the compressed \emph{centralized} approach. 
      Moreover, a naive algorithm that
      does not economize X-DC communication, as is the case of the
      \emph{distributed} method, also reduces transfers with respect
      to the current \emph{centralized} state-of-the-art.}
    \label{fig:transfer-datacenter}
\end{figure*}

\section{Results and Discussion}\label{sec:discussion}

In this section we present measurements from the methods introduced
above, using the datasets described in \OldS\ref{sec:data}.  We focus
on two key metrics: 1) total X-DC transfer size, and 2) latency to
model.

\subsection{Simulation}

\subsubsection{X-DC Transfer}

Figure \ref{fig:transfer-datacenter} illustrates the total X-DC
transfer of the different methods for different numbers of data
centers.  We only show two versions of CRITEO and WBCTR for space
limitations, though the others follow the same patterns.  In general,
X-DC transfers increase with the number of data centers as there are
more X-DC communication paths. As expected, increasing the model
dimensionality also impacts the transfers in the distributed versions.
In Figure \ref{fig:criteo-transfer-datacenter-100M}, the efficient
distributed approach (\emph{distributed-fadl}) performs at least 1
order of magnitude better than \emph{centralized} in every scenario,
achieving the biggest difference (2 orders of magnitude) for 2 data
centers. In this setting, \emph{centralized} (any variant) transfers
half of the compressed data (870 GB) through the X-DC link before
training, whereas \emph{distributed-fadl} just needs 9 GBs worth of
transfers to train the model.  Likewise, in the WBCTR dataset
(Figure~\ref{fig:wood-transfer-datacenter-160M}), we see the biggest
difference in the 2 data centers scenario (1 order of magnitude).
When the data is spread across 8 data centers, \emph{centralized}
transfers almost the same as \emph{distributed}.  In general, the non
communication-efficient \emph{distributed} baseline also outperforms
the current practice, \emph{centralized}, on both datasets.




\subsubsection{Loss / X-DC Transfer Trade-off}
\begin{figure*}[tbp!]
    \centering
    \begin{subfigure}{0.24\textwidth}
        \includegraphics[width=\textwidth]{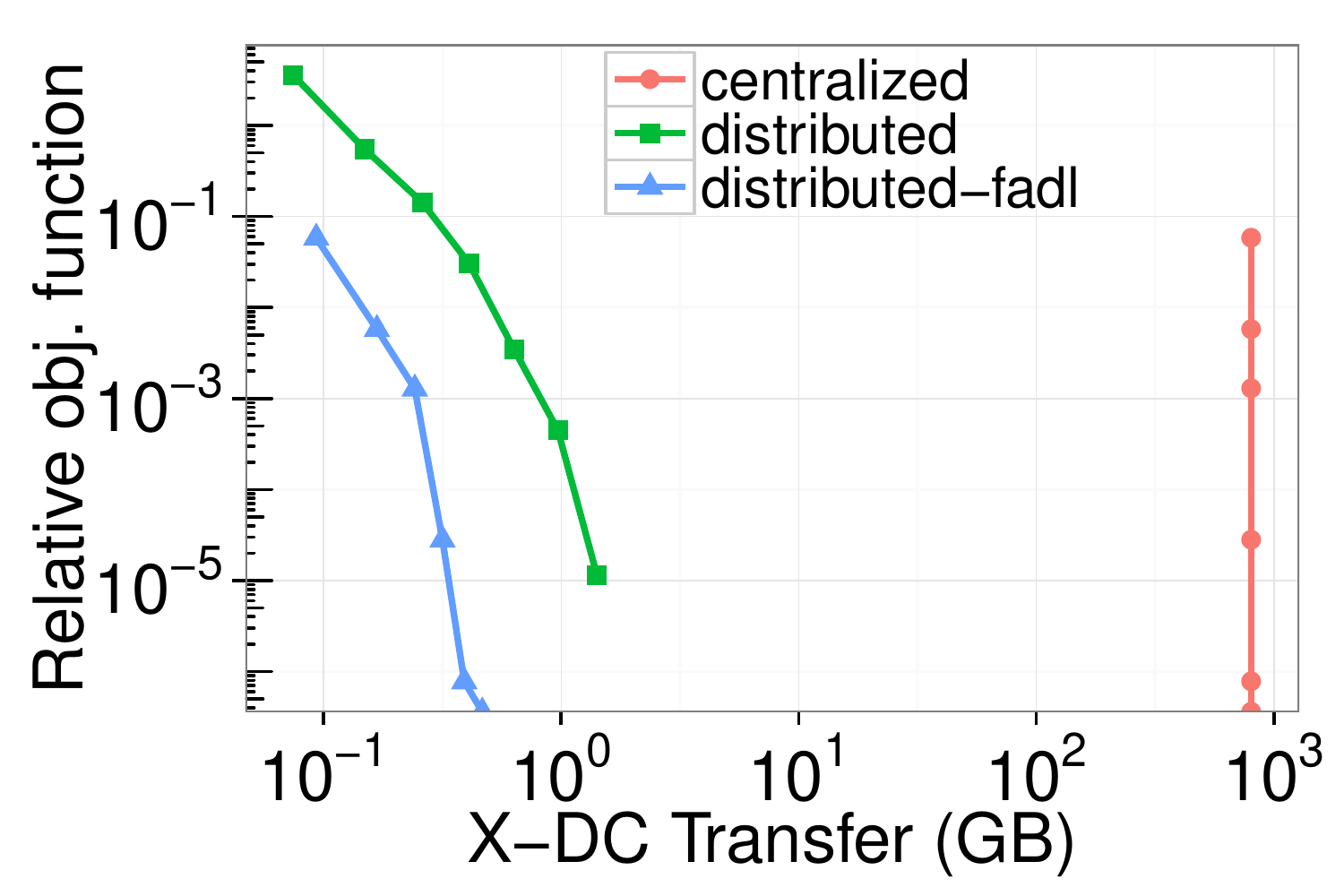}
        \caption{CRITEO 5M - 2DC} 
        \label{fig:criteo-loss-transfer-5M-2DC}
    \end{subfigure}
    \begin{subfigure}{0.24\textwidth}
        \includegraphics[width=\textwidth]{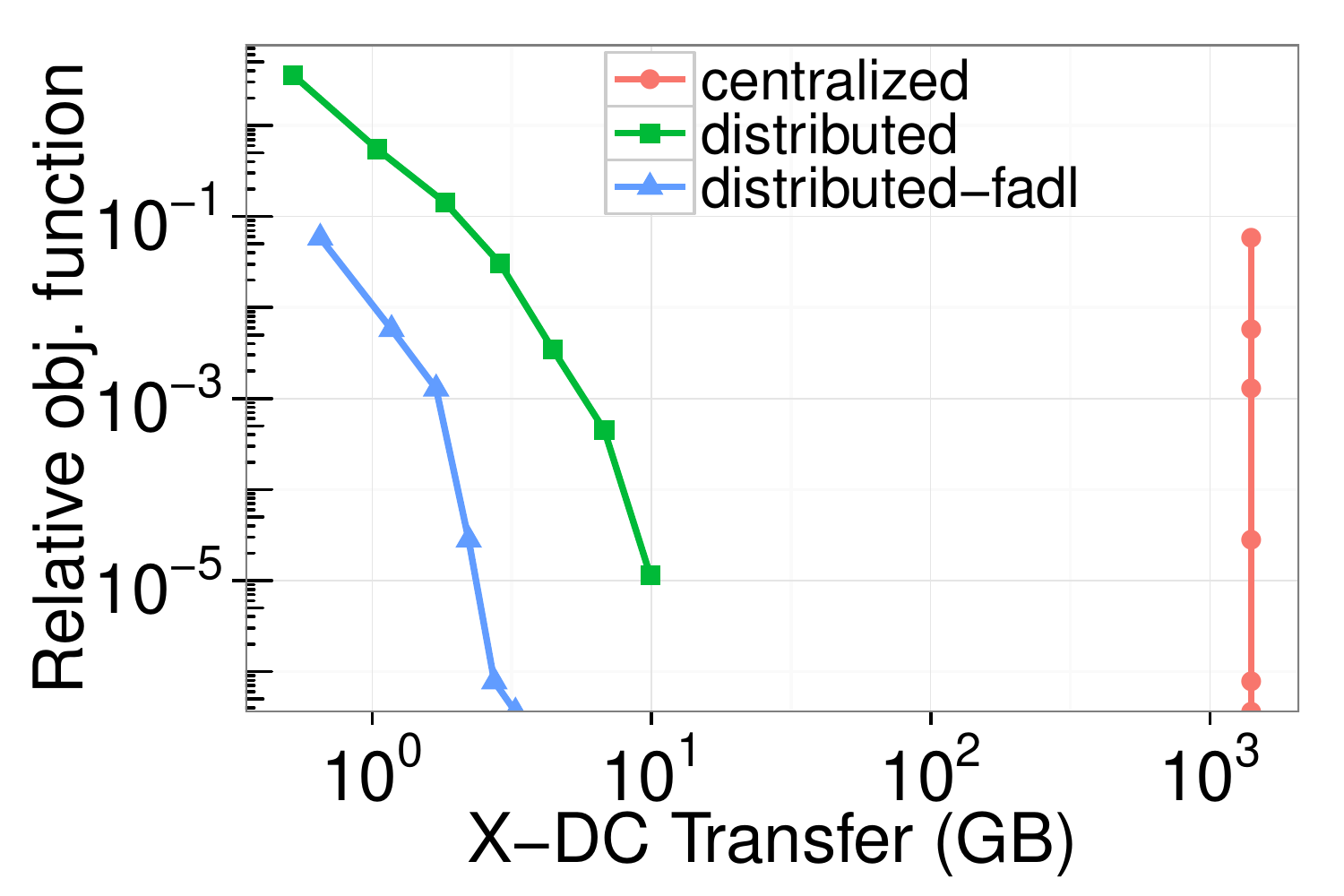}
        \caption{CRITEO 5M - 8DC}
        \label{fig:criteo-loss-transfer-5M-8DC}
    \end{subfigure}
    \begin{subfigure}{0.24\textwidth}
        \includegraphics[width=\textwidth]{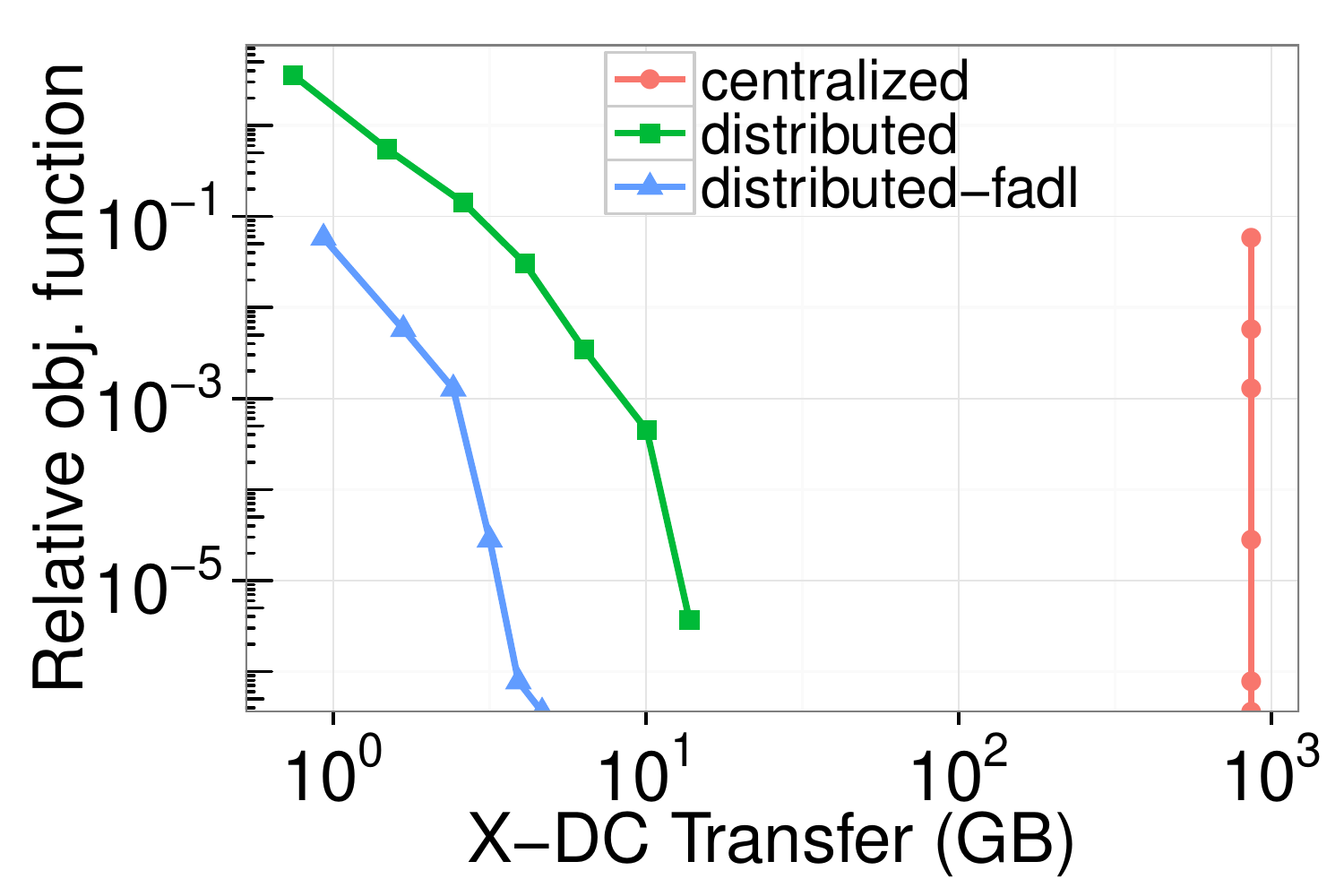}
        \caption{CRITEO 50M - 2DC} 
        \label{fig:criteo-loss-transfer-50M-2DC}
    \end{subfigure}
    \begin{subfigure}{0.24\textwidth}
        \includegraphics[width=\textwidth]{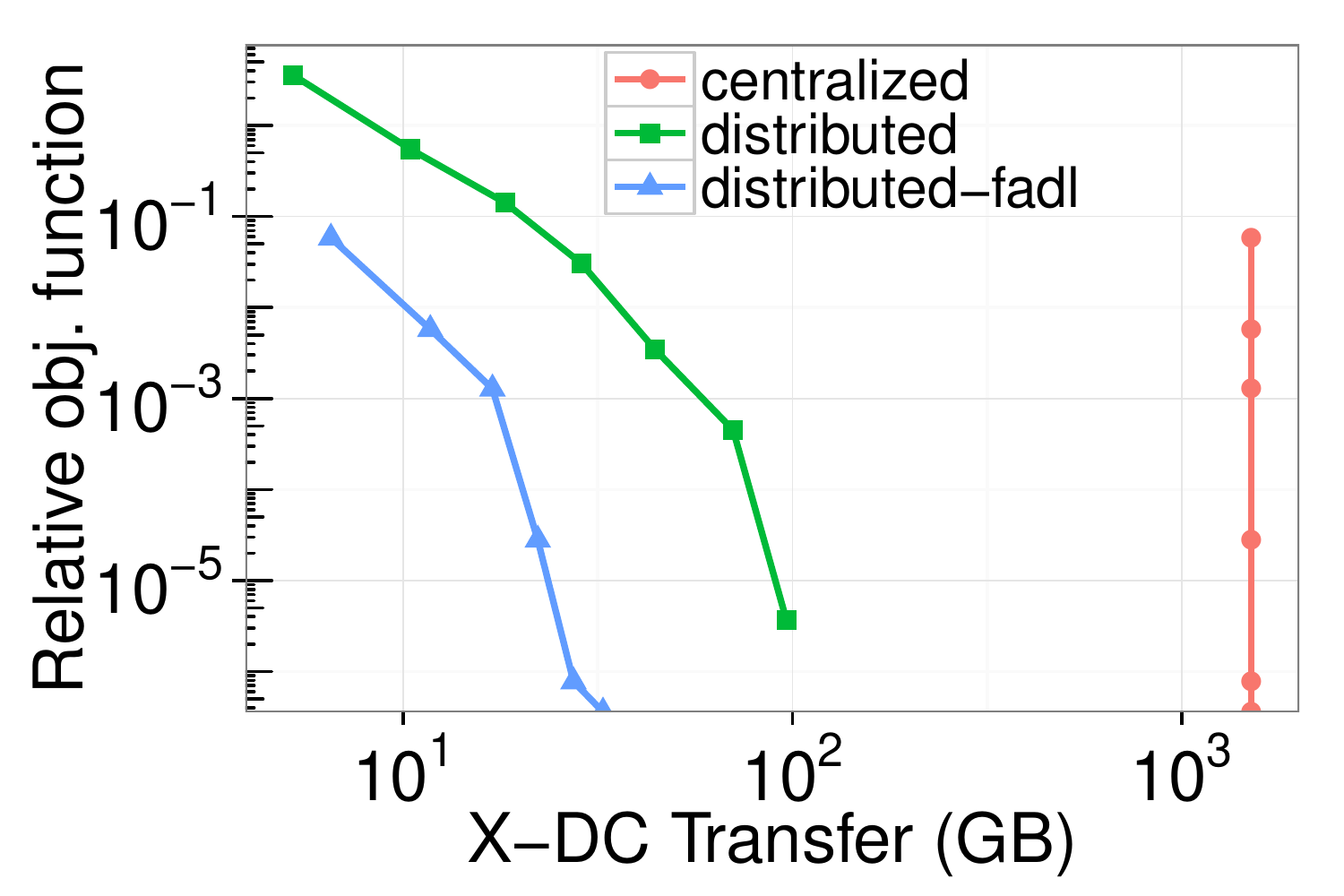}
        \caption{CRITEO 50M - 8DC}
        \label{fig:criteo-loss-transfer-50M-8DC}
    \end{subfigure}        
    \begin{subfigure}{0.24\textwidth}
        \includegraphics[width=\textwidth]{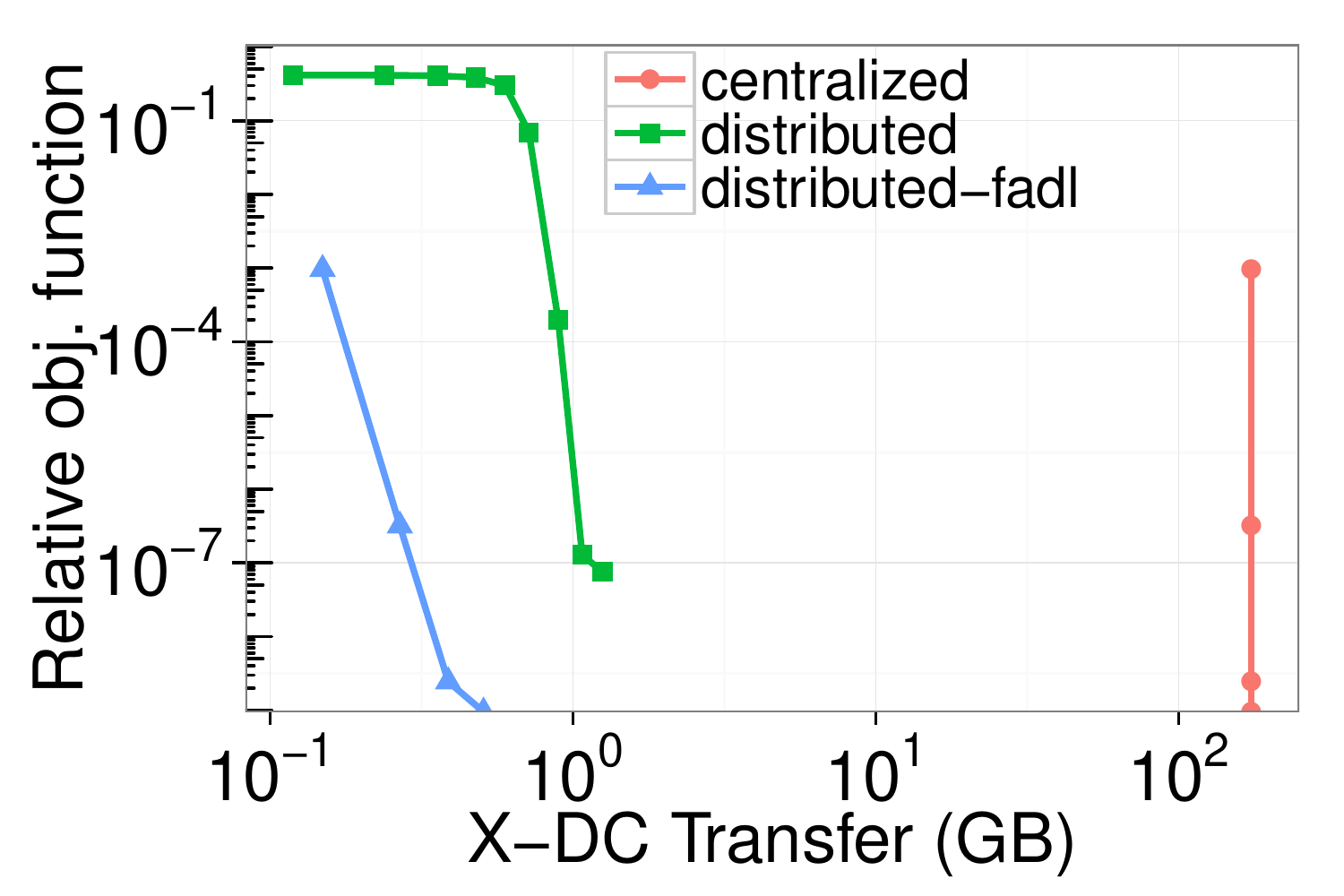}
        \caption{WBCTR 8M - 2DC} 
        \label{fig:wood-loss-transfer-8M-2DC}
    \end{subfigure}
    \begin{subfigure}{0.24\textwidth}
        \includegraphics[width=\textwidth]{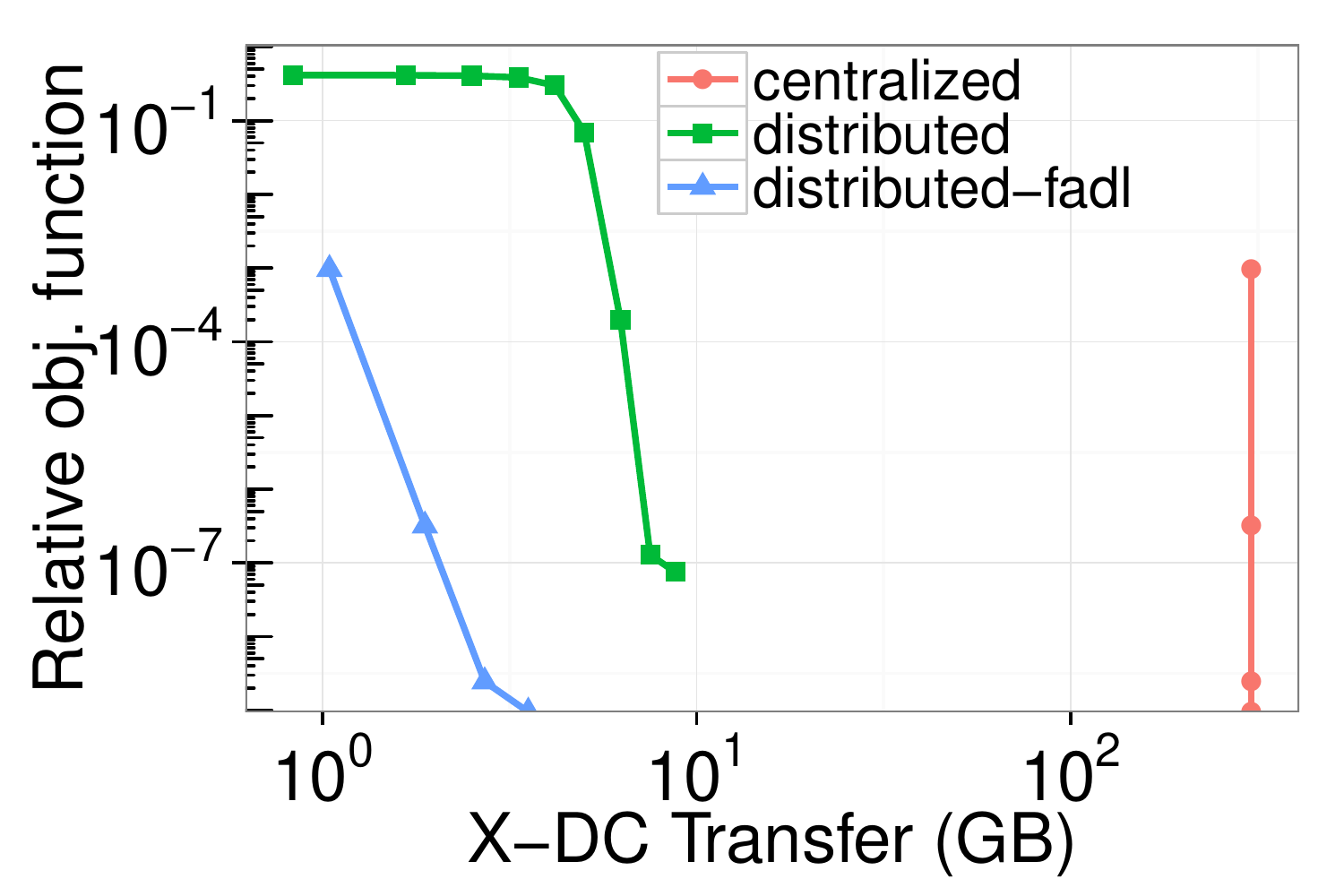}
        \caption{WBCTR 8M - 8DC}
        \label{fig:wood-loss-transfer-8M-8DC}
    \end{subfigure}
    \begin{subfigure}{0.24\textwidth}
        \includegraphics[width=\textwidth]{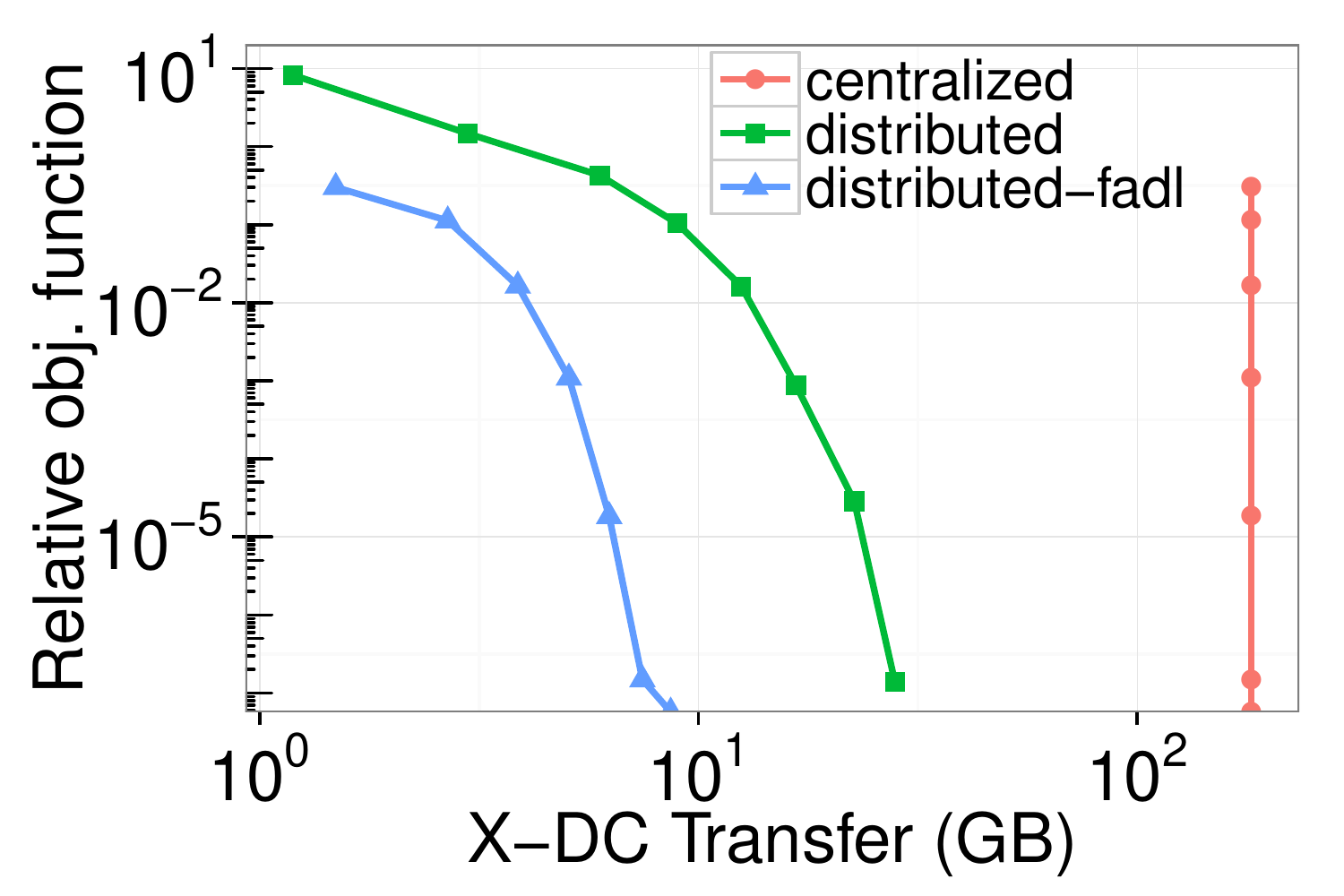}
        \caption{WBCTR 80M - 2DC} 
        \label{fig:wood-loss-transfer-80M-2DC}
    \end{subfigure}
    \begin{subfigure}{0.24\textwidth}
        \includegraphics[width=\textwidth]{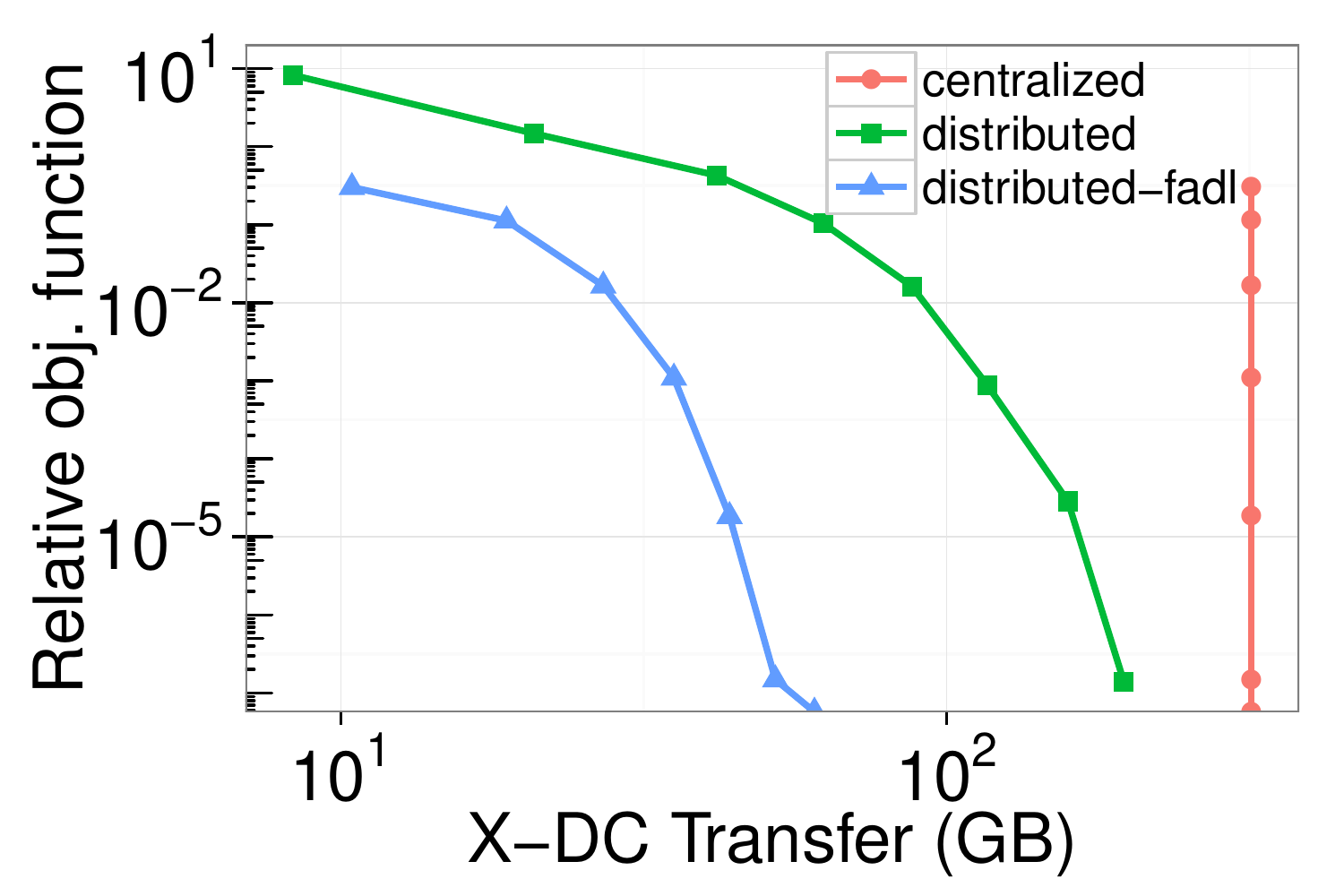}
        \caption{WBCTR 80M - 8DC}
        \label{fig:wood-loss-transfer-80M-8DC}
    \end{subfigure}
    \vspace{-0.1cm}
    \caption{Relative objective function (compared to the best) versus X-DC transfer (in GB) for 2 and 8 DCs for two versions of CRITEO and WBCTR datasets. The method \emph{distributed-fadl} achieves lower objective values much sooner in terms of X-DC transfers than the other methods. The \emph{centralized} objective remains constant with respect to X-DC transfers throughout the optimization as it starts once the data has been transferred. The \emph{distributed} method does incur in more transfers than \emph{distributed-fadl}, although it also reduces the overhead of the \emph{centralized} approach. Increasing the models dimensionality, naturally increases the X-DC transfers. Note that \emph{centralized} refers to both of its variants (\emph{stream} and \emph{bulk}), and we only report compressed data transfers for this method. }
    \label{fig:loss-transfer}
\end{figure*}

Commercial deployments of machine learning systems impose deadlines
and resource boundaries on the training process. This can make it
impossible to run the algorithm till convergence.  Hence, it is
interesting to study the performance of the centralized and
distributed approaches in relationship to their resource consumption.
Figure~\ref{fig:loss-transfer} shows the relative objective function
over time as a function of X-DC transfers for 2 and 8 data centers on
the CRITEO and WBCTR datasets. We use the relative difference to the
optimal function value, calculated as $(f-f^*) / f^*$, where $f^*$ is
the minimum value obtained across methods.  X-DC transfers remain
constant in the \emph{centralized} (any variant) method as it starts
the optimization after the data is copied, i.e. no X-DC transfers are
made while training.  In general, \emph{distributed-fadl} achieves
lower objective values much sooner in terms of X-DC transfers, which
means that this method can get some meaningful results faster. If an
accurate model is not needed (e.g. $10^{-2}$ relative objective
function value), \emph{distributed-fadl} gives a quicker response.  As
we increase the number of data centers, X-DC communication naturally
increases, which explains the right shift trend in the plots
(e.g. Figures \ref{fig:criteo-loss-transfer-5M-2DC} and
\ref{fig:criteo-loss-transfer-5M-8DC}).

\subsubsection{Storage}

As the number of data centers increases, \emph{centralized} (any variant) 
requires more space on disk. In particular, assuming the data 
is randomly partitioned across data centers, 
\emph{centralized} stores at least $1.5\times$ more data than 
the distributed versions, with a maximum difference of almost $2\times$
when considering 8 data centers.  On the other hand,
both \emph{distributed} and \emph{distributed-fadl} 
only need to store the original dataset
($1\times$) throughout the different configurations.

\begin{figure*}[tbp!]
    \centering
    \begin{subfigure}{0.30\textwidth}
        \includegraphics[width=\textwidth]{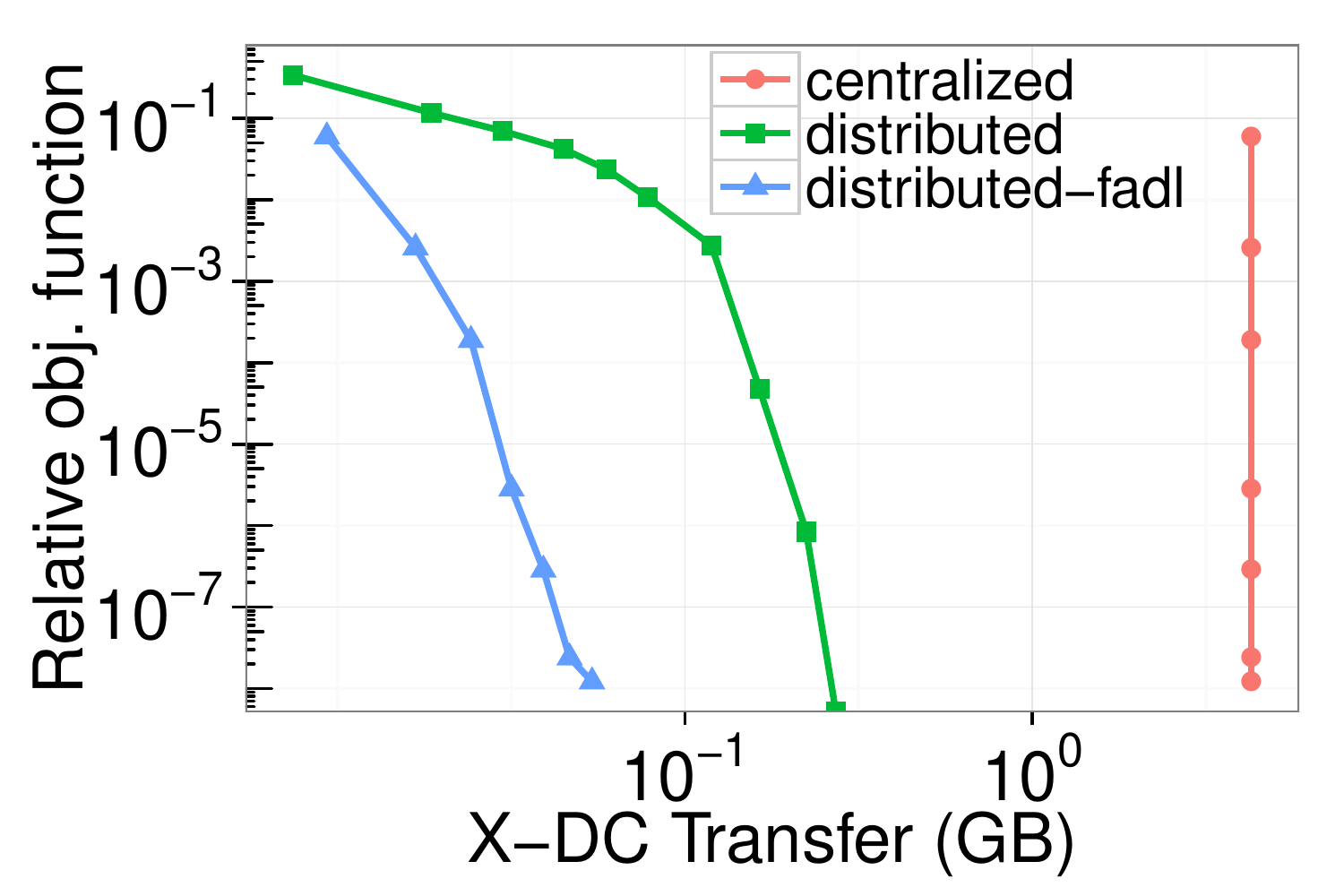}
        \caption{KAGGLE 500K} 
        \label{fig:kaggle-loss-transfer-500K}
    \end{subfigure}
    \begin{subfigure}{0.30\textwidth}
        \includegraphics[width=\textwidth]{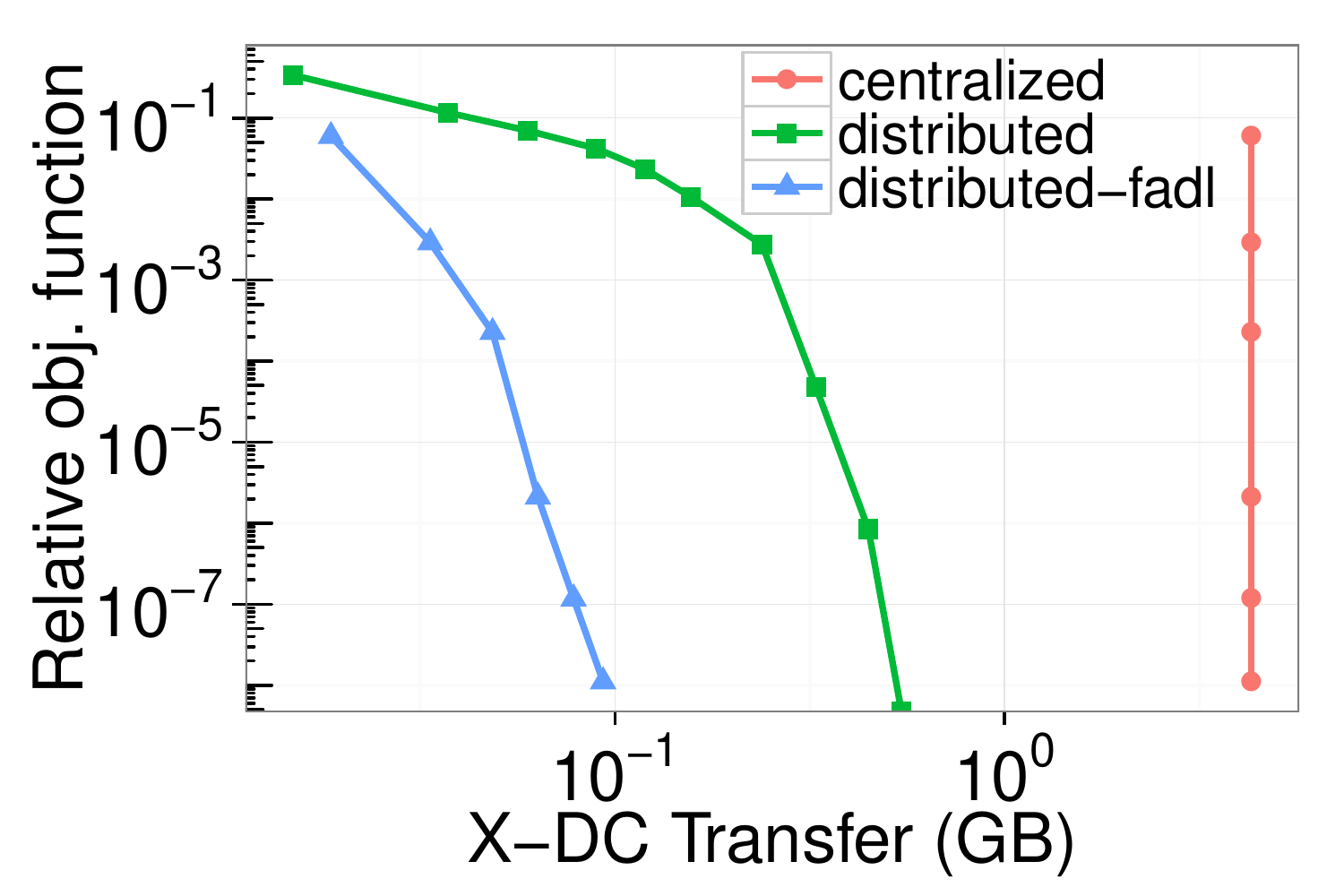}
        \caption{KAGGLE 1M}
        \label{fig:kaggle-loss-transfer-1M}
    \end{subfigure}
    \begin{subfigure}{0.30\textwidth}
        \includegraphics[width=\textwidth]{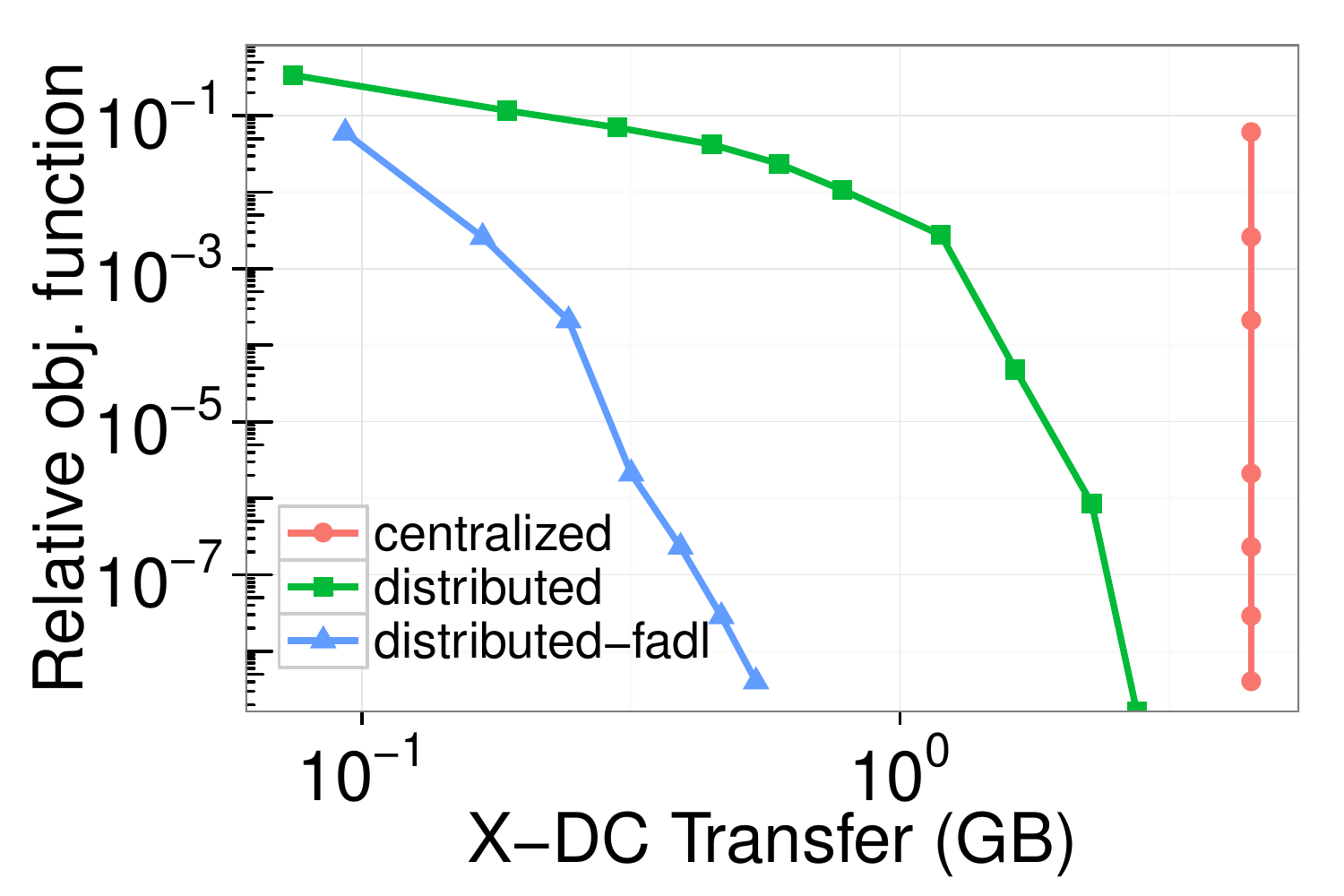}
        \caption{KAGGLE 5M}
        \label{fig:kaggle-loss-transfer-5M}
    \end{subfigure}
    \vspace{-0.1cm}
    \caption{Relative objective function (compared to the best) versus X-DC
      transfer (in GB) for the KAGGLE dataset in 2 Azure data centers. The
      increase in the model size explains the right shift trend in the
      plots. The method \emph{distributed-fadl} consumes the
      least amount of X-DC bandwidth, at least 1 order less in every scenario, and 2 when using 
      the 500K model. The loss/transfer pattern is similar to Figure
      \ref{fig:loss-transfer}. Both distributed methods transfer much less X-DC data than the \emph{centralized}
      state-of-the-art. Note that \emph{centralized} refers to both of its flavors (\emph{stream} and \emph{bulk}), and only transfers compressed data.}
    \label{fig:kaggle-loss-transfer}
\end{figure*}

\begin{figure*}[tbp!]
    \centering
    \begin{subfigure}{0.30\textwidth}
        \includegraphics[width=\textwidth]{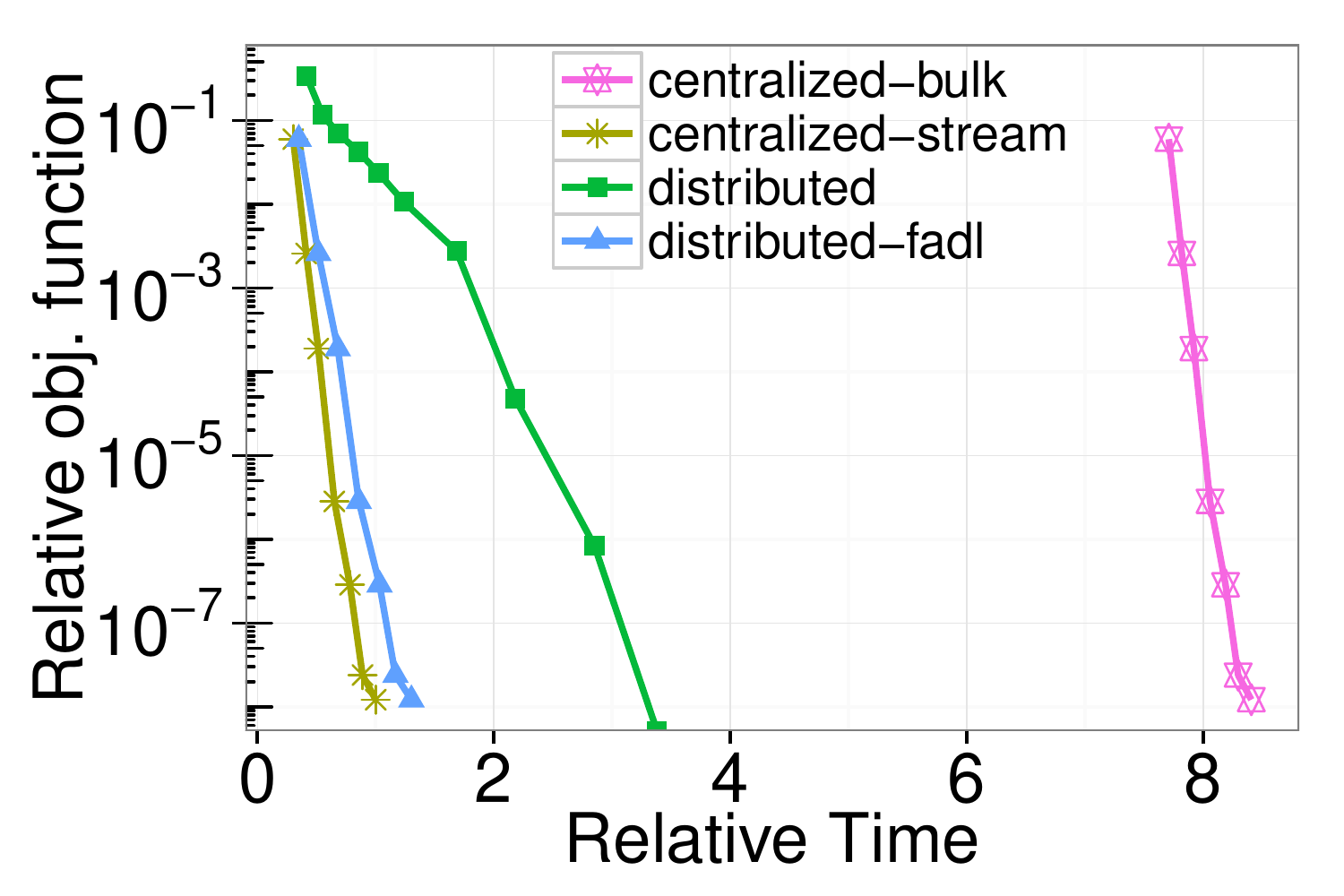}
        \caption{KAGGLE 500K} 
        \label{fig:kaggle-loss-time-500K}
    \end{subfigure}
    \begin{subfigure}{0.30\textwidth}
        \includegraphics[width=\textwidth]{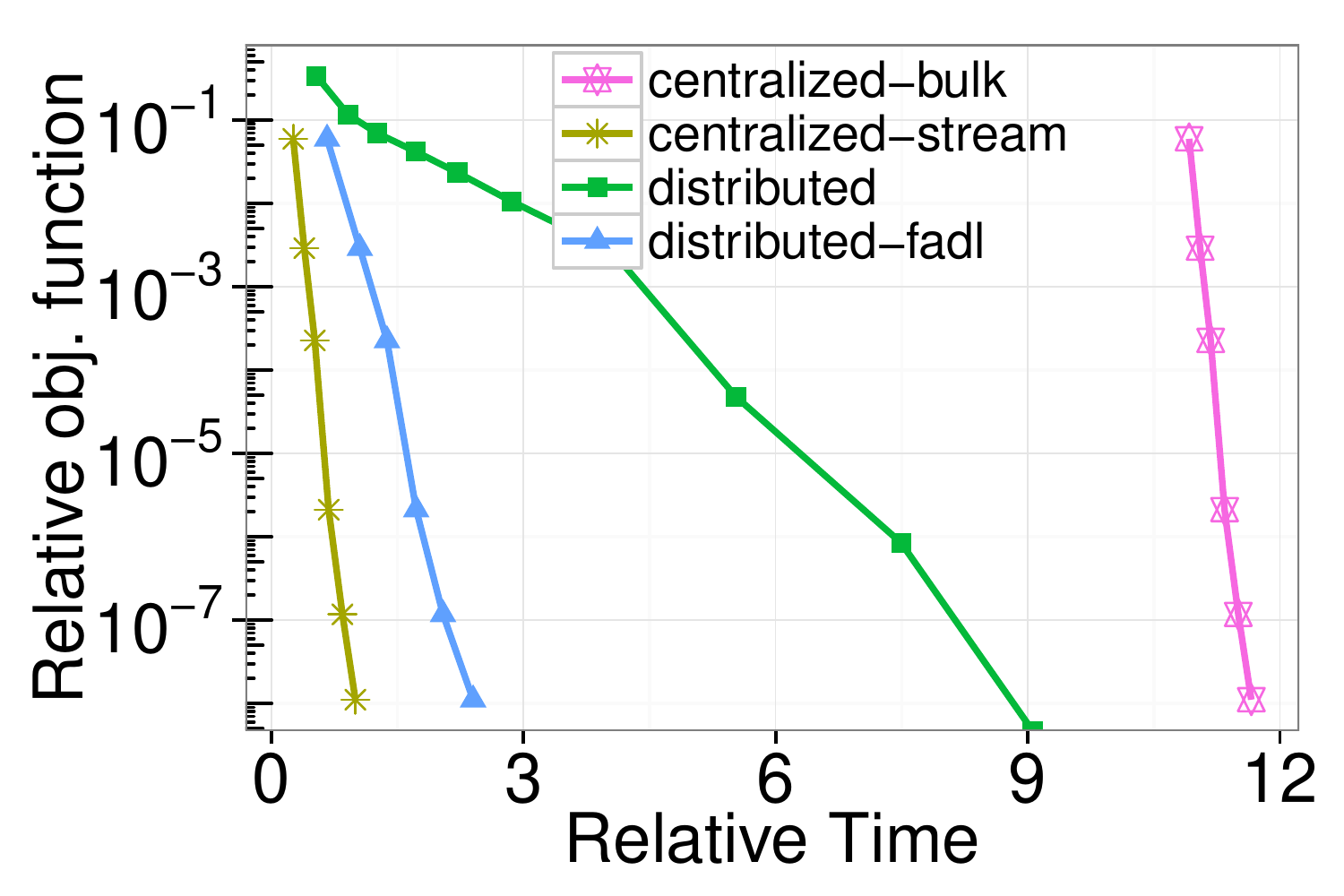}
        \caption{KAGGLE 1M}
        \label{fig:kaggle-loss-time-1M}
    \end{subfigure}
    \begin{subfigure}{0.30\textwidth}
        \includegraphics[width=\textwidth]{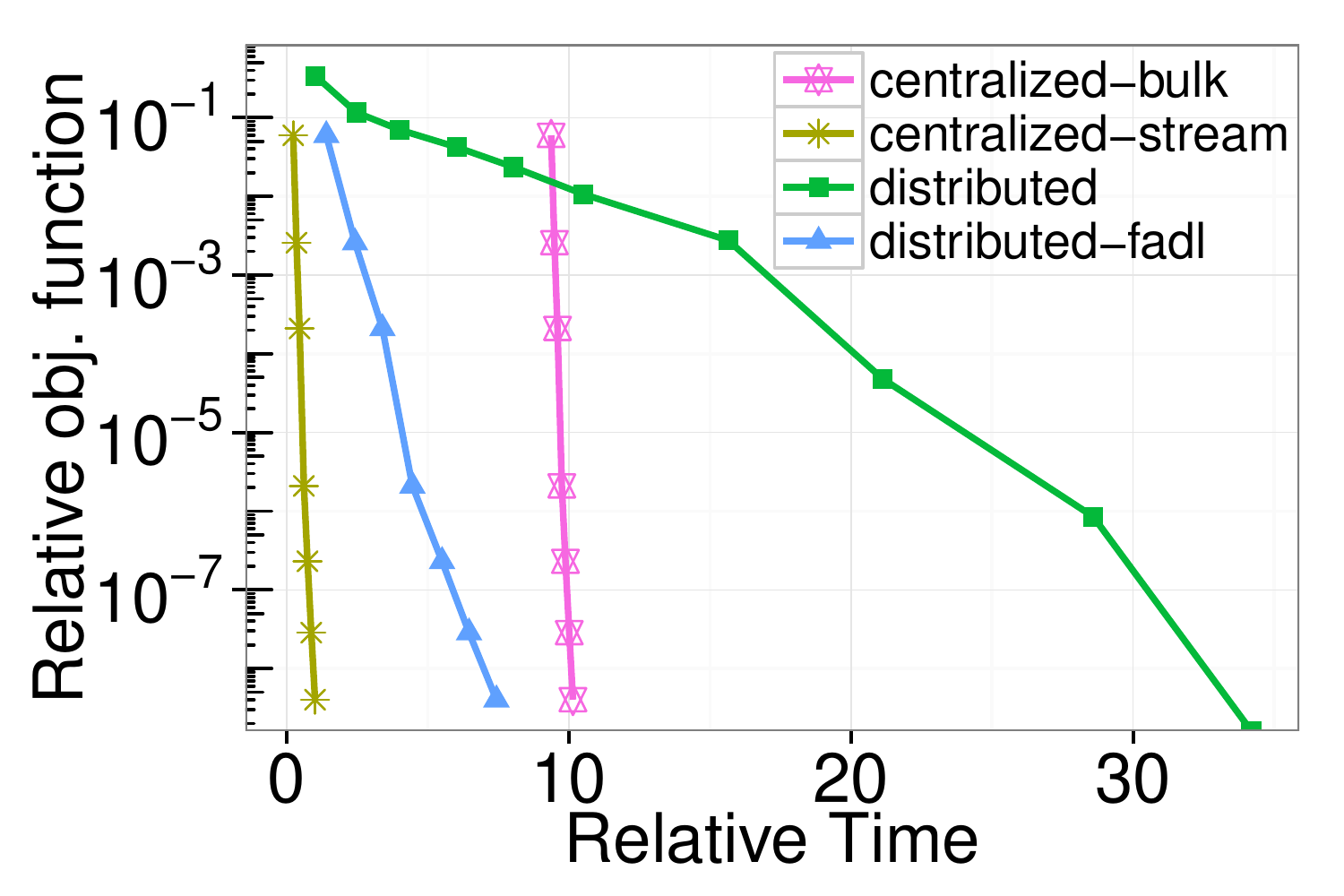}
        \caption{KAGGLE 5M} 
        \label{fig:kaggle-loss-time-5M}
    \end{subfigure}
    \vspace{-0.1cm}
    \caption{Relative objective function (compared to the best) over time
      (relative to the \emph{centralized-stream} method) for the KAGGLE
      dataset in 2 Azure data
      centers. The method \emph{distributed-fadl} beats every approach but
      \emph{centralized-stream}. This latter method is the best
      case scenario, where the data has already been copied and is
      available in a single data center when the job is
      executed. The \emph{distributed-fadl} method lies very close to the optimum 
      (\emph{centralized-stream}), especially in low-dimensional models and when considering
      commonly accepted objective function values ($10^{-4}$, $10^{-5}$).
      Both \emph{distributed} and \emph{distributed-fadl} 
      performance degrades when the model size increases (as expected), but \emph{distributed}
      does so much worse (\ref{fig:kaggle-loss-time-5M}), which further shows that in order to do
      geo-distributed machine learning, a communication-efficient algorithm is needed.}
      \label{fig:kaggle-loss-time}
\end{figure*}

\subsection{Real Deployment}

\subsubsection{X-DC Transfer}

To validate our simulation, we include
Figure~\ref{fig:kaggle-loss-transfer}, which shows the relative
objective function with respect to the X-DC bandwidth for the KAGGLE
dataset in 2 Azure DCs (Western US and Western Europe).  These
experiments completely match our findings in the simulated
environment. For the \emph{centralized} approach, we transfer the data
from EU to US, and run the optimization in the latter data center.
Similar to Figure \ref{fig:loss-transfer}, the increase in the number
of features expectedly causes more X-DC transfers (right shift trend
in the plots). The efficient geo-distributed method
\emph{distributed-fadl} still communicates the least amount of data,
almost 2 orders of magnitude less than the \emph{centralized} (any
variant) approach for the 500K model (Figure
\ref{fig:kaggle-loss-transfer-500K}).

\subsubsection{Runtime}

Figure \ref{fig:kaggle-loss-time} shows the relative objective
function over time for the 2 Azure data centers using the KAGGLE
dataset.  We choose the KAGGLE dataset because it is the most
challenging for our system in terms of impact on runtime. This is due
to the fact that the ratio of model-size/data-size is the largest
(i.e. proportionally larger model).  The cost of transferring the model at
each iteration impacts runtime more substantially---see discussion in
\OldS\ref{sec:problem}.  We normalize the time to the
\emph{centralized-stream} approach, calculated as $t / t^*$, where
$t^*$ is the overall time taken by \emph{centralized-stream}.  This method
performs the fastest in every version of the dataset (500K, 1M, and 5M
features) as the data has already been copied by the time it starts,
i.e. no copy time overhead is added, and represents the lower bound in
terms of running time.

Although the \emph{centralized} approach always transfers compressed data, we  
do not take into account the compression/decompression time for computing 
the \emph{centralized-bulk} runtime, which would have otherwise tied the results to 
the choice of the compression library.
Figures \ref{fig:kaggle-loss-time-500K}, \ref{fig:kaggle-loss-time-1M}, 
and \ref{fig:kaggle-loss-time-5M} show that \emph{centralized-bulk}
pays a high penalty for copying the data, it runs in approximately $8\times$ or more of 
its \emph{stream} counterpart.

The communication-efficient \emph{distributed-fadl} approach executes in 
$1.3\times$, $2.4\times$, and $7.4\times$ of the \emph{centralized-stream} baseline for 500K, 1M, and 5M
models respectively, which is a remarkable result given that it transfers orders of magnitude 
less data (Figure \ref{fig:kaggle-loss-transfer}), and executes in a truly 
geo-distributed manner, respecting potentially strict regulatory constraints. 
Moreover, if we consider the relative objective function values commonly used in practice to achieve 
accurate models ($10^{-4}$, $10^{-5}$), this method lies in the same ballpark as the lower bound 
\emph{centralized-stream} in terms of running time. Still, \emph{distributed-fadl} is way 
ahead in terms of X-DC transfers (orders of magnitude of savings in X-DC bandwidth), 
while at the same time it potentially complies with data sovereignty regulations.

Figure \ref{fig:kaggle-loss-time-500K} shows that
\emph{distributed-fadl} performs very close to the best scenario,
matching the intuition built in \OldS\ref{sec:problem} that this
method does very well on tasks with (relatively) small models and
(relatively) large number of examples. Furthermore, this efficient
method also runs faster than \emph{distributed} in every setting,
which further highlights the importance and benefits of the algorithm
introduced in \OldS\ref{sec:algo}.

Finally, \emph{distributed} performance degrades considerably as the
model size increases.  In particular, this method does a poor job when
running with 5M features (Figure \ref{fig:kaggle-loss-time-5M}), which
concurs with the intuition behind the state-of-the-art
\emph{centralized} approach: copying the data offsets the
communication-intensive nature of (naive) machine learning
algorithms. We see that this intuition does not hold true for the
efficient algorithm described in \OldS\ref{sec:algo}.

\section{Related Work}\label{sec:related_work}

Prior work on systems that deal with geographically distributed
datasets exists in the literature.  The work done by Vulimiri et
al. \cite{Vulimiri:2015:WAN,Vulimiri:2015:GA} poses the thesis that
increasing global data and scarce WAN bandwidth, coupled with
regulatory concerns, will derail large companies from executing
centralized analytics processes. They propose a system that supports
SQL queries for doing X-DC analytics. Unlike our work, they do not target iterative machine
learning workflows, neither focus on jobs latency. They mainly care about reducing 
X-DC data transfer volume.

Pu et al. \cite{Pu:2015:SIGCOMM_DATA} proposes a
low-latency distributed analytics system called Iridium. Similar to
Vulimiri et al., they focus on pure data analytics and not on machine learning
tasks. Another key
difference is that Iridium optimizes task and data placement across
sites to minimize query response time, while our system respects stricter 
sovereignty constraints and does not move raw data around.

JetStream is a system for wide-area streaming data analysis that
performs efficient queries on data stored ``near the edge''. 
They provide different approximation techniques to 
reduce the data size transfers at the expense of accuracy. One of such techniques is 
dropping some fraction of the data via sampling. 
Similar to our system, they only move \emph{important} data to a centralized 
location for global aggregation (in our case, we only move gradients and models), and
they compute local aggregations per site prior to sending (in our case, we perform local optimizations per data center using the algorithm described in \OldS\ref{sec:algo}).

Other existing Big Data processing systems, such as Parameter Server, 
Graphlab, or Spark \cite{Li:2014:PS, Low:2012:GRAPHLAB, Zaharia:2012:Spark}, 
efficiently process data in the context of a single data center, which typically employs a high-bandwidth network. To the best of our knowledge, they have not been tested in multi-data center deployments (and were not designed for it), where scarce WAN bandwidth makes it impractical to naively communicate parameters between locations. Instead, our system was specifically optimized to perform well on this X-DC setting.

Besides the systems solutions, the design of efficient distributed machine learning algorithms has also been the topic of a broad research agenda. 
The Terascale method \cite{JMLR:v15:agarwal14a} might be the best representative
method from the statistical query model class and is considered a state-of-the-art solver. 
CoCoA \cite{DBLP:conf/nips/JaggiSTTKHJ14} represents the class of distributed dual methods that, in
each outer iteration, solve (in parallel) several local dual optimization problems.
Alternating Direction Method of Multipliers (ADMM) \cite{Boyd:2011:DOS:2185815.2185816, DBLP:journals/jmlr/ZhangLS12} is a dual method different from the primal method we use here, however, it also solves approximate problems in the nodes and iteratively reaches the full batch solution.
Recent follow up work \cite{Dhruv:2015:FADL} shows that the algorithm described in section \OldS\ref{sec:algo} performs better than the aforementioned ones, both in terms of communication passes and running time.





\section{Open Problems}\label{sec:openproblems}

GDML is an interesting, challenging and open area of research.
Although we have proposed an initial and novel geo-distributed
approach that shows substantial gains over the centralized
state-of-the-art in many practical settings, many open questions remain, both from a systems as
well as a machine learning perspective.

Perhaps, the most crucial aspect is fault tolerance.  With data
centers distributed across continents, consistent network connectivity
is harder to ensure than within a single data center, and network
partitioning is more likely to occur.  On other hand, a DC-level
failure might completely compromise the centralized approach (if the
primary DC is down), while the geo-distributed solution might continue
to operate on the remaining data partitions.  There has been some
initial work~\cite{Shravan:2014:RAML} to make ML algorithms tolerant
to missing data (e.g. machine failures). This work assumes randomly
distributed data across partitions, hence, a failure removes an
unbiased fraction of the data.  In production settings, this is the
case when multi-DC deployments are created for load-balancing
(e.g. within a region)---we are aware of multiple such scenarios
within Microsoft infrastructure.  However, cross-region deployments
are often dictated by latency-to-end-user considerations.  In such
settings, losing a DC means losing a heavily biased portion of the
population (e.g. all users residing in Western US).  Coping with
faults and tolerating transient or persistent data unavailability is
an open problem that will likely require both system and ML
contributions.

In this work we have restricted ourselves to linear models with $l_2$
regularization, and shown results on logistic regression models. It
would be interesting to validate similar observations in other
regularizers (e.g. $l_1$).  More broadly, studying geo-distributed
solutions that can minimize X-DC transfers for other complex learning
problems such as kernels, deep-nets, etc., is still an open area of
research.

Lastly, a truly geo-distributed approach surely does no worse than a
centralized method when analyzed from regulatory and data sovereignty
angles.  Questions in this area arise not only at the global scale,
where different jurisdictions might not allow raw data sharing, but
also at the very small scale, e.g. between data stored in a private
cluster and data shared in the cloud.  We believe that studying the
setup presented here from a privacy-preserving and
regulatory-compliance angle will yield important improvements, and
potentially inform regulators.

Besides presenting early results in this area, this paper is intended
as an open invitation to researchers and practitioners from both
systems and ML communities.  We foresee the need for substantial
advances in theory, algorithms and system design, as well as the
engineering of a whole new practice of Geo-Distributed Machine
Learning (GDML).  To that end, we contributed back all the changes
done to Apache Hadoop YARN and Apache REEF as part of this work.



\section{Conclusions}\label{sec:conclusions}
Large organizations have a planetary footprint with users scattered in
all continents. Latency considerations and regulatory requirements
motivate them to build data centers all around the world. From this, a
new class of learning problems emerge, where global learning tasks
need to be performed on data ``born'' in geographically disparate data
centers, which we call Geo-Distributed Machine Learning (GDML).

To the best of our knowledge, this aspect of machine learning has not
been studied in the literature before, despite being faced by
practitioners on a daily basis.

In this work, we introduce and formalize this problem, and challenge
common assumptions and practices.  Our empirical results show that a
geo-distributed system, combined with communication-parsimonious
algorithms can deliver a substantial reduction in costly and scarce
cross data center bandwidth.  Further, we speculate distributed
solutions are structurally better positioned to cope with the quickly
evolving regulatory frameworks.

To conclude, we acknowledge this work is just a first step of a long
journey, which will require significant advancements in theory,
algorithms and systems.

\bibliography{gdml}{}
\bibliographystyle{unsrt}

\end{document}